\documentclass{article}

\PassOptionsToPackage{numbers,sort&compress}{natbib}
\usepackage[final,main]{neurips_2026}

\usepackage[utf8]{inputenc}
\usepackage[T1]{fontenc}
\usepackage{hyperref}
\usepackage{url}
\usepackage{booktabs}
\usepackage{amsfonts}
\usepackage{amsmath}
\usepackage{amssymb}
\usepackage{graphicx}
\usepackage{nicefrac}
\usepackage{microtype}
\usepackage{xcolor}
\usepackage{makecell}
\usepackage{multirow}
\usepackage{subcaption}

\definecolor{AnchorBlue}{HTML}{0072B2}
\definecolor{StressOrange}{HTML}{D55E00}
\definecolor{BoostGreen}{HTML}{009E73}
\definecolor{EmbedGray}{HTML}{7A7A7A}

\title{Where to Look Is Not How to Fix:\\
Pre-Denoising Diagnostics and Modality-Dependent Control\\
in Diffusion Composition}

\author{%
  Fangzheng Wu \\
  Department of Computer Science\\
  Tulane University\\
  New Orleans, LA 70118\\
  \texttt{fwu6@tulane.edu}
  \And
  Brian Summa \\
  Department of Computer Science\\
  Tulane University\\
  New Orleans, LA 70118\\
  \texttt{bsumma@tulane.edu}
}

\hypersetup{
  pdftitle={Where to Look Is Not How to Fix: Pre-Denoising Diagnostics and Modality-Dependent Control in Diffusion Composition},
  pdfauthor={Fangzheng Wu and Brian Summa},
  pdfsubject={Diagnostic accessibility and intervention responses in diffusion models},
  pdfkeywords={diffusion models, compositionality, diagnostic accessibility, cross-attention}
}
\usepackage{placeins}
\begin{document}
\raggedbottom

\maketitle

\begin{abstract}
Understanding compositional failures in text-to-image diffusion
requires identifying both where stress is detectable and how
intervention changes the output. We study these questions
through a controlled anchor--stress protocol that jointly
evaluates text-encoder diagnostics and denoiser interventions.
We introduce a text-only Compositional Stress Index (CSI),
which separates common from rare compositions across SD1.5,
SDXL, and the SD3 text path and provides an upstream
diagnostic coordinate. A matched six-prompt localization
study links intervention location to distinct outcomes:
residual-minimizing embedding adapters improve
representation fit, while downstream cross-attention
intervention increases color hit rate (CHR) by 0.0272.
Across SD1.5 and SDXL denoiser blocks, the largest positive
signed diagnostic-accessibility mean occurs at the deep
encoder, whereas selective boost has its largest positive
mean CHR response at decoder blocks. Selective subtraction
and broad ablation reveal further modality- and
architecture-dependent responses, including a substantial
CHR decrease when SDXL decoder cross-attention is broadly
ablated.
We find a diagnosis-control dissociation under our
controlled attribute-object composition setting:
compositional defects are diagnosable before denoising,
but the representation coordinate that exposes risk is
not necessarily the coordinate or modality that improves
generation.
\end{abstract}

\section{Introduction}
\label{sec:intro}

Text-to-image diffusion models~\citep{ho2020ddpm,rombach2022ldm,podell2024sdxl} have achieved remarkable generation quality but remain brittle on semantically rare attribute--object compositions~\citep{huang2023t2icompbench,yuksekgonul2023aro}: a model that reliably generates ``a red car'' may fail on ``a blue pizza'' or ``a glass cat,'' producing the wrong color, material, or texture. Two threads of work address this. Inference-time methods---Attend-and-Excite~\citep{chefer2023attend}, SynGen~\citep{rassin2024syngen}, StructureDiffusion~\citep{feng2023structureddiffusion}, Divide-and-Bind~\citep{li2023divideandbind}, and attention-map-controlled synthesis~\citep{wang2023compositional}---modify denoiser cross-attention so that attribute tokens better bind to the correct object regions, typically aggregating corrections across all layers. A second line of work studies the upstream text-conditioning pathway, asking whether compositional failures originate in representation geometry~\citep{lewis2024clipbind,bhalla2024splice} or in misalignment between the embedding objective and the denoiser's binding readout~\citep{zarei2024wiclp,marioriyad2024attention}. Implicit in both threads is an assumption: the pipeline location where a compositional defect is most \emph{detectable} is also the location where it is most effectively \emph{correctable}. We test this relationship by comparing diagnostic readouts and intervention responses at distinct pipeline sites.

Our central finding is a site--modality dissociation: compositional stress has an upstream diagnostic signature, while intervention responses depend jointly on pipeline location and the type of edit. In the tested UNets, the deep encoder has the largest positive signed diagnostic-accessibility mean, whereas decoder blocks have the largest positive mean response to selective boost. This motivates treating diagnosis and intervention as distinct empirical questions: \emph{where to look is not how to fix}.

To make this concrete, consider modern text-to-image diffusion pipelines~\citep{rombach2022ldm,ramesh2022hierarchical,saharia2022photorealistic,podell2024sdxl,esser2024scaling}. A text encoder first produces conditioning embeddings, which are then consumed by a denoising backbone to guide generation. In the UNet-based Stable Diffusion family, including SD1.5 and SDXL~\citep{rombach2022ldm,podell2024sdxl}, these conditioning embeddings are injected through cross-attention layers distributed across encoder, middle, and decoder blocks~\citep{rombach2022ldm,hertz2022prompt,tumanyan2023plug}. This architecture provides multiple internal sites at which a compositional failure may be readable, and multiple sites at which an intervention may affect the output. 
We use this UNet hierarchy for our block-level localization experiments, while also testing the representation-side diagnostic on an additional non-UNet text-to-image pipeline, namely Stable Diffusion 3~\citep{esser2024scaling}.
Our question is therefore: at which sites can compositional failures be \emph{detected}, and at which can they be \emph{fixed}, and does detection at a site imply fixability there?

Table~\ref{tab:teaser} maps this dissociation across three pipeline sites; rows correspond to sites, columns distinguish diagnostic readout from intervention outcomes. We present the evidence in two parts, corresponding to two regions of the pipeline.

\textbf{Upstream: diagnosis and repair at the text encoder} (Table~\ref{tab:teaser}, row~1).
If a text encoder processes ``a blue pizza'' compositionally, the full-prompt representation should be roughly recoverable from the separate representations of ``blue,'' ``pizza,'' and the surrounding context. A large deviation from this component-based reconstruction indicates that the full prompt is not factorizing into its parts. We call this deviation a \emph{non-factorization residual} and aggregate it into a Compositional Stress Index (CSI), where higher CSI indicates stronger departure from compositional factorization. Within a deliberately controlled anchor--stress setting that pairs common and rare attribute--object compositions while holding the object and context fixed, CSI separates the two groups perfectly (area under the receiver operating characteristic curve, ROC-AUC${=}1.0$) across SD1.5, SDXL, and the SD3 text path, and provides a diagnostic coordinate for the subsequent intervention analyses. We use the text-only CSI component for the pre-denoising interpretation (Appendix~\ref{app:csi}).

This diagnostic coordinate lets us examine how representation changes relate to output changes. We train low-rank adapters to move stress embeddings toward anchor-like structure and compare their responses with denoiser-side cross-attention intervention. The adapters improve representation fit. On a matched six-prompt localization subset, downstream intervention increases CHR by 0.0272, while embedding intervention decreases it. The comparison connects intervention location to distinct representation and image-level outcomes under the tested objectives (\S\ref{sec:results2}).

\textbf{Downstream: per-block modality decomposition inside the denoiser} (Table~\ref{tab:teaser}, rows~2--3).
This representation--output comparison motivates the downstream question: once text conditioning is read out by the denoiser, which internal sites are diagnostic, and which are causally controllable? For UNet-based SD1.5 and SDXL, we compare block groups spanning the deep encoder, middle block, and decoder, using three intervention modalities: selective boosting, selective zeroing, and broad cross-attention ablation. 
We quantify denoiser-side readout by diagnostic accessibility (DA), a block-level measure of how well cross-attention patterns separate anchor from stress prompts during baseline generation. 
The results reveal a second dissociation, now \emph{within} the denoiser: D2 has the largest positive signed DA mean, while decoder blocks have the largest positive mean CHR response to selective boosting. Broad ablation produces a different, architecture-dependent profile, with a substantial CHR decrease at SDXL U0 and at early SD1.5 encoder blocks. These profiles characterize how intervention responses vary with both \emph{where} an edit is applied and \emph{how} it modifies cross-attention (\S\ref{sec:results3}).

\paragraph{Contributions.}
(1)~Under the tested intervention families, we uncover a diagnosis--control dissociation in text-to-image diffusion: the site where compositional stress is most readable is not necessarily the site where intervention is most effective.
(2)~We further decompose this dissociation under the tested intervention families, showing that selective token edits and broad cross-attention ablations have different site profiles across deep encoder and decoder blocks.
(3)~We introduce text-only CSI as a pre-denoising stress coordinate and evaluate it across SD1.5, SDXL, and the SD3 text path under the controlled anchor--stress protocol.
(4)~Under the tested intervention families, we connect representation and output responses through a matched localization study: residual-minimizing adapters improve embedding fit, while downstream intervention increases CHR on the selected six-prompt subset.

\begin{table}[t]
\centering
\caption{Diagnostic readouts and observed intervention responses across pipeline sites. Entries summarize measured profiles rather than significance decisions; effect sizes, uncertainty, and statistical tests are reported in Sections~\ref{sec:results2}--\ref{sec:results3}. DA peaks refer to signed means.}
\label{tab:teaser}
\small
\begin{tabular}{@{}l@{\;\;}c@{\;\;}c@{\;\;}c@{}}
\toprule
Pipeline site & Diagnostic readout & Targeted edits & Broad ablation \\
\midrule
Text encoder & Residual signal & Residual reduction & --- \\
Deep encoder & Positive DA peak & Small CHR changes & Architecture-dependent \\
Decoder & Smaller signed DA & Peak mean boost CE & SDXL U0: decrease \\
\bottomrule
\end{tabular}
\end{table}
\section{Related Work}
\label{sec:related}

The cross-attention-steering methods introduced in \S\ref{sec:intro}~\citep{chefer2023attend,rassin2024syngen,feng2023structureddiffusion,li2023divideandbind,wang2023compositional} share a design choice: they aggregate attention signal across all layers and resolutions, treating cross-attention as a homogeneous control surface; logit-level masking and entity-region constraints (e.g., MaskDiffusion-style methods, box-guided attention control) target specific resolutions or layers but similarly do not decompose across stages. WiCLP~\citep{zarei2024wiclp} pivots upstream, showing embedding-level repair \emph{can} work with supervision aligned to the denoiser's readout and implicating objective misalignment rather than the embedding space itself. Our findings should not be interpreted as evidence that all upstream interventions fail. WiCLP, for instance, demonstrates that embedding-space interventions can improve compositional generation when optimized with denoising-aligned objectives. Our result is narrower: reducing the CSI-diagnosed residual via residual minimization is not sufficient to guarantee semantic improvement. \citet{marioriyad2024attention} link entity-missing to attention overlap. A complementary line of work intervenes at training or sampling time rather than at inference attention, including contrastive curriculum fine-tuning~\citep{han2025progressive}, sampling-time corrective steering~\citep{dutta2025co3}, and PAC-Bayesian priors over attention distributions~\citep{jiang2024pacbayes}. Our contribution is orthogonal to all of these: we diagnose \emph{where} in the pipeline compositional risk appears and characterize \emph{which modality} works at each site, producing a layer-selection principle that could inform any of these approaches---training-time methods might focus supervision on the deep-encoder representation we identify, while decoder-side attention is a candidate target for evaluating sampling-time correctors.

On the representation and evaluation side, CLIP embeddings admit sparse linear decompositions~\citep{bhalla2024splice} and compose single-object attributes but fail under multi-object binding~\citep{lewis2024clipbind}. Prompt-to-Prompt~\citep{hertz2022prompt} showed cross-attention encodes spatial and semantic structure; FreeU~\citep{si2024freeu} demonstrated that UNet encoder--decoder information flow is a control axis. More broadly, probing and causal intervention research~\citep{alain2017understanding,belinkov2022probing,vig2020causal,geiger2021causal,meng2022locating} has shown that probe-friendly layers need not be intervention-friendly. We sharpen this gap for diffusion cross-attention by identifying both \emph{which layer} and \emph{which modality} works at each layer, and we quantify a pre-denoising non-factorization signal not captured by these prior analyses. Output-side benchmarks~\citep{huang2023t2icompbench,thrush2022winoground,li2022blip} measure \emph{whether} composition fails; our work diagnoses \emph{where} the failure originates and \emph{how} to intervene at each site. Extended related work covering compositional evaluation suites, object-centric representations, and energy-based compositional generation appears in Appendix~\ref{app:related_extended}.

\section{Setup: Sites, Diagnostics, and Interventions}
\label{sec:setup}

We study controlled compositional stress in text-to-image diffusion. Each example consists of an \emph{anchor} prompt paired with a single \emph{stress} prompt that shares the same object and context but differs in the target attribute. The anchor uses a common attribute--object composition, whereas the stress prompt uses a rarer or less typical composition. Given such pairs, we ask two questions: where in the model is the stress signal diagnostically readable, and where can an intervention causally improve the generated output?

Our experiments instantiate this setup with 432 prompts organized into 216 paired groups, each consisting of one anchor prompt and one stress prompt. Attribute--object combinations are selected from T2I-CompBench~\citep{huang2023t2icompbench}, CLEVR~\citep{johnson2017clevr}, and ARO~\citep{yuksekgonul2023aro} lexicons to span common and rare compositional regimes (examples in Appendix~\ref{app:prompts}). For example, within one group the anchor might be ``a studio photo of a red pizza'' and the stress prompt ``a studio photo of a blue pizza''---same object, same context, different attribute. Examples such as ``a red car'' and ``a blue pizza'' that appear elsewhere in the paper illustrate the rarity gradient across the dataset, not within-pair variation.

Primary models for generation and denoiser-side intervention are SD1.5 and SDXL, while SD3 is used only for text-path diagnostic extension. For UNet block-group analysis, we partition cross-attention modules by their position in the UNet architecture. Following standard notation, $D_k$ denotes the $k$-th down-sampling (encoder) stage, $M$ the middle bottleneck, and $U_k$ the $k$-th up-sampling (decoder) stage. SDXL has five active cross-attention groups: $\{D_1, D_2, M, U_0, U_1\}$; SD1.5 has seven: $\{D_0, D_1, D_2, M, U_1, U_2, U_3\}$ (module counts per group in Appendix~\ref{app:blocks}). Throughout, we refer to $D_2$ as the \emph{deep encoder} and $U_0$ (SDXL) / $U_1$ (SD1.5) as the \emph{decoder}. With these sites defined, we next introduce two diagnostic readouts---one at the text encoder, one inside the denoiser---and two families of intervention: embedding-level repair and denoiser-side cross-attention edits.

\paragraph{Text-encoder diagnostic: CSI.}
We operationalize text-side compositionality as a component-reconstruction test. If the text encoder preserves a factorized representation of attribute, object, and context, then a simple composition rule should explain the full-prompt embedding; large residuals indicate prompt-specific interactions beyond this factorized account. For each prompt $p_{\mathrm{full}}$ (e.g., ``a studio photo of a blue pizza''), we construct component inputs: $p_{\mathrm{attr}}$ (the target attribute, e.g., ``blue''), $p_{\mathrm{obj}}$ (the object, e.g., ``pizza''), and $p_{\mathrm{ctx}}$ (the surrounding context template). Let $h(\cdot)$ denote the representation extracted from the text encoder, and let $\hat{h}(\cdot)$ be a deliberately low-capacity composition model fit to predict $h(p_{\mathrm{full}})$ from the component representations. We consider additive, low-rank linear, and low-rank bilinear variants, fit on anchor prompts and evaluated on both anchor and stress prompts. We define the non-factorization residual as
\begin{equation}
R(p_{\mathrm{full}}) =
\frac{
\left\lVert h(p_{\mathrm{full}}) -
\hat{h}(p_{\mathrm{attr}}, p_{\mathrm{obj}}, p_{\mathrm{ctx}})
\right\rVert_2
}{
\left\lVert h(p_{\mathrm{full}}) \right\rVert_2 + \varepsilon
}.
\end{equation}
where $\varepsilon$ prevents division by zero.
In plain terms: If the encoder preserves component-wise structure for ``blue,'' ``pizza,'' and the context template, then a simple composition model should be able to reconstruct the full-prompt embedding, yielding a small $R$. If the full prompt instead induces prompt-specific interactions that are not predictable from those components, the residual increases.
CSI aggregates this normalized residual with span-geometry features via robust standardization on anchor statistics; we report $\mathrm{CSI}_{\mathrm{text}}$, $\mathrm{CSI}_{\mathrm{span}}$, and combined $\mathrm{CSI}$. Only $\mathrm{CSI}_{\mathrm{text}}$ is strictly pre-denoising; combined CSI includes internal attention-derived span features. Encoder extraction, calibration scope, and span-feature definitions are given in Appendix~\ref{app:csi}.

\paragraph{Denoiser diagnostic: DA.}
The second diagnostic readout operates inside the denoiser. For each UNet block group $b$, diagnostic accessibility $\mathrm{DA}(b)$ summarizes the signed difference in attribute-binding attention between matched anchor and stress prompts during baseline generation. We define an attribute-binding measure $\mathrm{ABM}_b$ as the fraction of attribute-token attention mass falling within top-quantile object-attention regions, then compute:
\begin{equation}
\mathrm{DA}(b)
=
\mathbb{E}_{(p_a,\,p_s)}
\bigl[
\mathrm{ABM}_b(p_a) - \mathrm{ABM}_b(p_s)
\bigr],
\end{equation}
where $(p_a, p_s)$ denotes a matched anchor--stress pair. Positive DA means anchor prompts show tighter attribute-to-object binding at block $b$ than stress prompts. We rank sites by this signed mean difference; negative DA indicates the reverse contrast and is not evidence of absent diagnostic information. Significance against zero does not establish a significant difference between sites. Formal definitions and validation of ABM are given in Appendix~\ref{app:abm}.

To separate \emph{where} an intervention is applied from \emph{what kind} of intervention it is, we organize denoiser-side edits along two axes: \emph{selectivity}, whether the edit targets only the attribute token or the entire cross-attention map, and \emph{sign}, whether the edit strengthens or weakens the conditioning channel. This gives a $2{\times}2$ design; the broad-additive cell is omitted because uniformly scaling all attention weights is canceled by post-softmax normalization, leaving three testable denoiser-side families. \emph{Selective additive} (\textsc{boost}) amplifies attention to the target attribute token at a chosen block, testing whether stronger local attribute readout improves binding. \emph{Selective subtractive} (\textsc{attr\_zero}) removes the same attribute-token attention, testing the opposite direction as a matched-selectivity control. \emph{Broad subtractive} (\textsc{zero}) uniformly flattens all cross-attention at a chosen block, testing whether the block's entire text-conditioning contribution is necessary or harmful. Together with two embedding-level repair families---pooled and token low-rank adapters trained to move stress representations toward anchor-like structure---this yields five intervention families. Intervention settings are given in Appendix~\ref{app:modality}; statistical testing is described below.

These three denoiser modalities span a natural hypothesis space about the defect's structure. If the failure reflects weak attribute readout, \textsc{boost} should help by amplifying the target-attribute channel. If the same channel is causally involved but harmful or mislocalized, \textsc{attr\_zero} tests the opposite direction by removing it. If the failure is distributed across the block's entire text-conditioning pathway, \textsc{zero} tests whether removing that contribution improves generation. Testing all three at each block characterizes how intervention responses depend jointly on site and modality. All intervention hyperparameters are fixed across block comparisons (strength, mask quantile, timestep schedule, and CFG settings in Appendix~\ref{app:modality}).

We evaluate control with object-local color hit rate (CHR): the fraction of pixels inside a fixed target-object mask whose color matches the desired attribute. The object mask is derived from baseline final-timestep attention and held fixed across conditions, which fixes the scoring region but does not independently verify object identity or localization. Scoring details are in Appendix~\ref{app:chr}, and external-mask sensitivity checks are in Appendix~\ref{app:color_sensitivity}. In the full per-block runs, CHR is finite for 52 stress prompts per seed; the other 164 stress prompts have no target color recorded by the scorer. This is a scoring-eligibility criterion, not a verified taxonomy of prompt semantics. We use chromatic binding because it changes attribute semantics while largely preserving object shape and localization, providing the cleanest endpoint for site-localization comparisons. Color prototypes and thresholds are reported in Appendix~\ref{app:chr}. For the per-block results, we first average paired differences across seeds, then test across prompts (CE) or matched \texttt{group\_id} units (DA). We use two-sided sign-flip tests (5,000 Monte Carlo draws, plus-one correction; exact enumeration for at most 16 units) and 2,000 percentile-bootstrap draws, with analysis seed 0. Reported 95\% confidence intervals are pointwise, not simultaneous intervals adjusted for multiple comparisons. DA is measured as an ABM difference and CE as a CHR difference; their magnitudes are not directly comparable. BH correction is applied separately to DA and CE across all active blocks within each model/run. Full CE analyses use 52 prompts with five seeds for boost/attr-zero and three for broad zeroing; DA uses 216 groups. Group IDs do not by themselves establish independence of shared anchor texts.

\section{Result 1: Text Encoders Provide a Pre-Denoising Stress Axis}
\label{sec:results1}

\begin{figure}[t]
\centering
\includegraphics[width=0.8\linewidth]{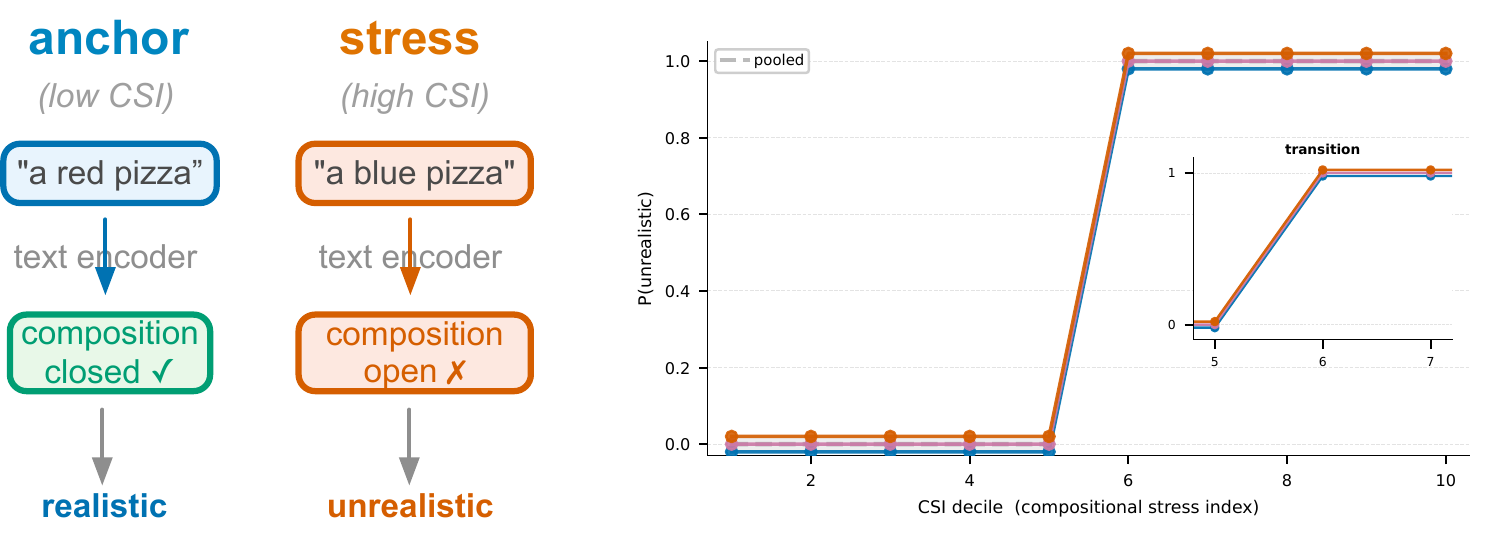}
\caption{Cross-model CSI decile curves. Probability of stress membership increases monotonically with CSI decile; SD1.5, SDXL, and SD3 collapse to the same transition pattern. ROC-AUC${=}1.0$ in all three models.}
\label{fig:csi_discovery}
\end{figure}

We first ask whether compositional stress is already visible before denoising, using only text-encoder representations. The goal is not to establish CSI as a general-purpose detector of arbitrary compositional failure, but to obtain a calibrated axis that orders prompts by stress severity before testing where intervention can change the generated output. On the controlled anchor--stress protocol, CSI cleanly separates common-composition anchors from rare-composition stress prompts across SD1.5, SDXL, and the SD3 text path (ROC-AUC${=}1.0$, pair-ranking${=}1.0$; Figure~\ref{fig:csi_discovery}). This ceiling-level separation should be interpreted with the protocol in mind: stress prompts are deliberately constructed to be semantically rare, and a frequency-only baseline already reaches ROC-AUC${=}0.79$. Thus, the important claim is not that CSI is a universal binary detector, but that text-side non-factorization provides a stable pre-denoising stress coordinate.

\begin{figure}[t] 
\centering 
\includegraphics[width=0.8\linewidth]{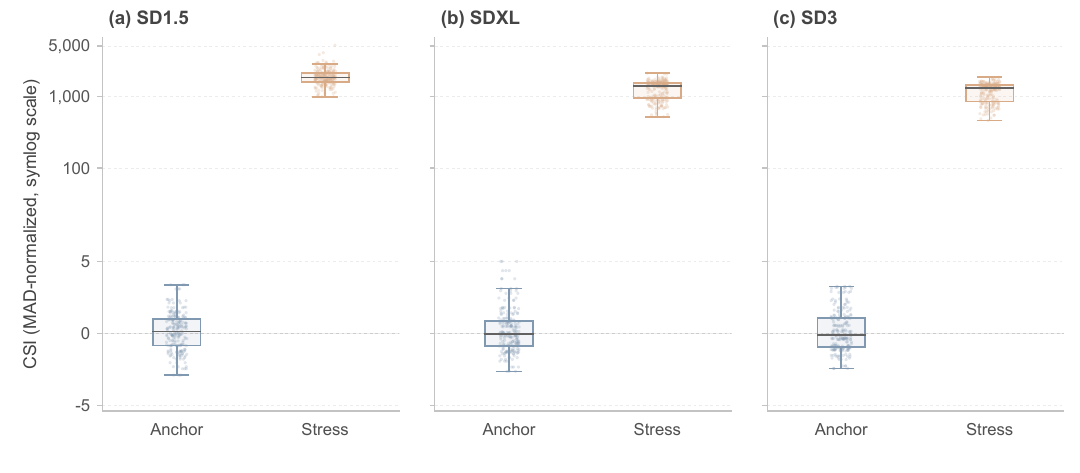} 
\caption{ Prompt-level CSI distributions for anchor and stress prompts across SD1.5, SDXL, and SD3. Values are displayed on a MAD-normalized symlog scale. Although absolute magnitudes are not directly comparable across models, stress prompts consistently occupy a much higher CSI regime than anchor prompts. This visualization complements ROC-AUC by showing the magnitude of the anchor–stress separation.} 
\label{fig:csi_distribution} 
\end{figure} 

While ROC-AUC summarizes ranking performance, Figure~\ref{fig:csi_distribution} visualizes the underlying prompt-level distributions. Across all three models, stress prompts exhibit substantially larger CSI values than anchors, consistent with the near-perfect separation reported above.

The stronger evidence is that CSI remains informative \emph{within} the stress set and aligns with output-side failures. Across stress prompts, CSI correlates with graded stress severity (decile-stress Spearman $\rho{=}0.87$) and with BLIP-derived attribute failure rate (Spearman $\rho{=}0.68$); output-side BLIP-VQA labels ($n{=}4320$) show that CSI predicts attribute failure with ROC-AUC${=}0.89$ and binding failure with ROC-AUC${=}0.85$. Controls support this interpretation: group-preserving permutation gives null ROC-AUC $=0.49 \pm 0.01$ (all empirical $p{=}0.001$), text-side non-factorization is the dominant contributor, span-only geometry is substantially weaker, and robustness to lexical confounds and family ablations is reported in Appendix~\ref{app:csi}. The pattern is also consistent across encoder families: SD1.5 uses a single CLIP text encoder, SDXL uses dual encoders (CLIP-L + OpenCLIP-G), and the SD3 text-path extension uses a T5-XXL sequence mean. Despite these architectural differences, the non-factorization gap appears in all three settings, so we use CSI as a model-agnostic pre-denoising stratification axis for the intervention analyses in \S\ref{sec:results2}--\S\ref{sec:results3}.

\section{Result 2: Upstream Representation Repair Does Not Imply Semantic Repair}
\label{sec:results2}

The upstream analysis established that compositional stress is detectable at the text encoder. We next ask whether repairing that diagnostic coordinate is sufficient to improve generated output. A pooled adapter reduces the representation residual ($\Delta_\text{repr}{=}-0.100$, $q{=}0.0024$, 20 prompts), and a token adapter gives $\Delta_\text{repr}{=}-0.184$ ($q{=}0.0008$, 19 prompts). Each has seven CHR-evaluable prompts, with observed $\Delta_\text{CHR}{=}0$ ($q{=}1$). On a separate CSI-stratified localization subset with six CHR-evaluable prompts and five seeds each, downstream intervention increases CHR by $+0.0272$ (95\% CI $[0.0092,0.0469]$; exact prompt-level $p{=}0.03125$, $q{=}0.046875$, $d_z{=}1.04$), whereas token-level embedding repair decreases CHR by $-0.0680$ (95\% CI $[-0.1071,-0.0327]$; $p{=}0.03125$, $q{=}0.03125$). The two arms use the same six prompts and seeds and have identical baseline CHR values. These results concern the measured CHR endpoint and do not establish general semantic repair.

A separate rank sweep (4--128) shows no significant positive CHR effect for the pooled adapters and negative CHR effects for the token adapters under within-run correction (Appendix~\ref{app:embed}). Residual improvement is not monotonic in rank, and these data do not rule out capacity or objective choice as explanations. In the localization downstream arm, early ABM increases ($\Delta{=}0.000450$, $q{=}0.0012$), whereas the mid-ABM change does not reach significance ($\Delta{=}0.000257$, $q{=}0.0708$). This pattern motivates, but does not identify, distinct readout mechanisms.

We interpret this as a representation--generation boundary: the upstream coordinate that diagnoses compositional stress is not automatically a control coordinate for the denoiser's binding behavior. This conclusion is specific to residual-minimizing objectives; it does not rule out embedding-level repair with binding-aligned supervision such as WiCLP~\citep{zarei2024wiclp}. The next question is whether the binding channel can be reached more directly inside the denoiser, and if so, at which blocks and with which modality. \S\ref{sec:results3} addresses this with a per-block sweep of cross-attention interventions.

\section{Result 3: Diagnostic and Control Sites Dissociate Inside the Denoiser}
\label{sec:results3}

Having found no positive CHR improvement from the tested residual-minimizing adapters, we next intervene inside the denoiser. D2 has the largest positive signed DA mean in both models across the baseline datasets (Appendix~\ref{app:da_stability}). Under boost baselines, SDXL has $\mathrm{DA}(\mathrm{D2}){=}+0.0039$ ($q{=}0.0174$), and SD1.5 has $+0.0282$ ($q{=}0.00028$). These tests establish a positive anchor--stress contrast at D2; they do not test whether D2 is significantly more diagnostic than every other block.

\begin{figure}[t]
\centering
\includegraphics[width=\linewidth]{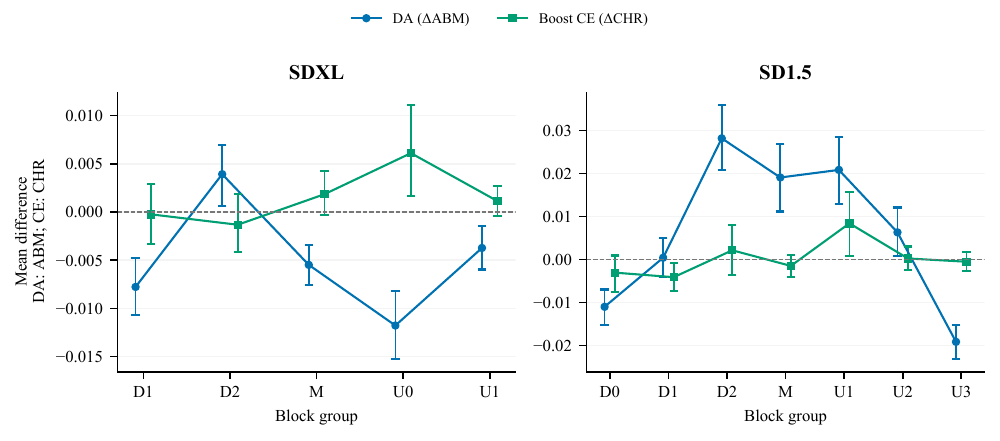}
\caption{Signed diagnostic accessibility (DA, blue) and selective boost CE (green) across active cross-attention block groups. DA is the anchor-minus-stress difference in ABM; CE is the intervention-minus-baseline difference in CHR. Both are dimensionless, but measure different quantities, so their magnitudes are not directly comparable. DA uses 216 matched anchor--stress groups; CE uses 52 evaluable stress prompts. Paired differences are averaged over five seeds before inference. Error bars show pointwise 95\% percentile-bootstrap confidence intervals (2,000 resamples). D2 has the largest positive DA mean, whereas U0 (SDXL) and U1 (SD1.5) have the largest positive boost CE means. No boost block survives within-model BH correction. See Appendices~\ref{app:da_stability} and~\ref{app:full_tables}.}
\label{fig:da_ce}
\end{figure}

\begin{figure}[t]
\centering
\includegraphics[width=\linewidth]{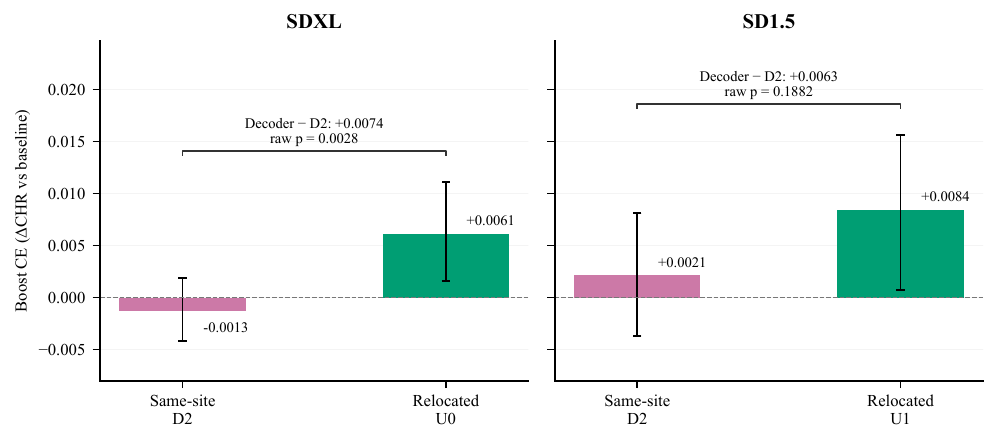}
\caption{Same-site (D2, purple) and relocated (decoder, green) boost effects relative to baseline. Error bars show pointwise 95\% percentile-bootstrap confidence intervals (2,000 resamples). Each estimate uses 52 prompt-level differences, averaged over five paired seeds. Brackets report direct decoder-minus-D2 contrasts and unadjusted two-sided sign-flip $p$-values: SDXL $+0.0074$ ($p=0.0028$); SD1.5 $+0.0063$ ($p=0.1882$). Decoder-versus-baseline BH $q$-values are 0.0840 and 0.1344, respectively. See Appendix~\ref{app:samesite}.}
\label{fig:samesite}
\end{figure}

\begin{figure}[t]
\centering
\setlength{\tabcolsep}{1pt}
\renewcommand{\arraystretch}{0.5}
\small
\begin{tabular}{@{}c@{\;}c@{\;}c@{\;}c@{\;}c@{}}
& Anchor & Stress baseline & D2 boost & Decoder boost \\[2pt]
\raisebox{0.8cm}{\rotatebox{90}{\parbox{1.6cm}{\centering\scriptsize SDXL\\``gold guitar''}}} &
\includegraphics[width=0.22\linewidth]{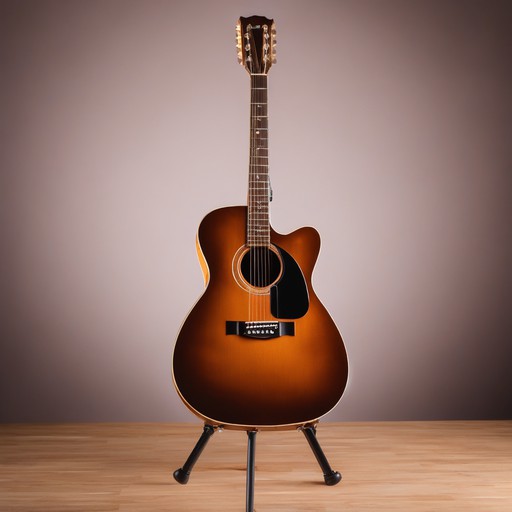} &
\includegraphics[width=0.22\linewidth]{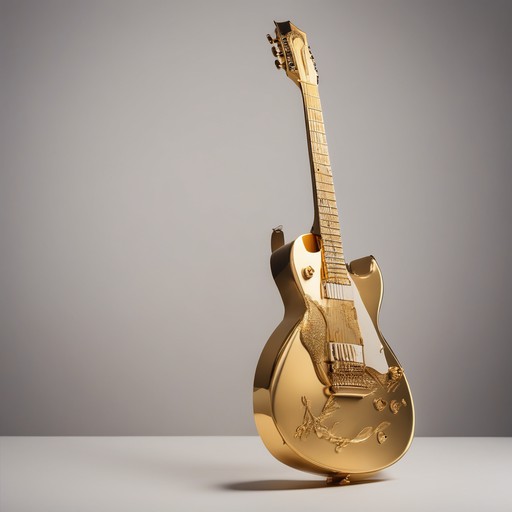} &
\includegraphics[width=0.22\linewidth]{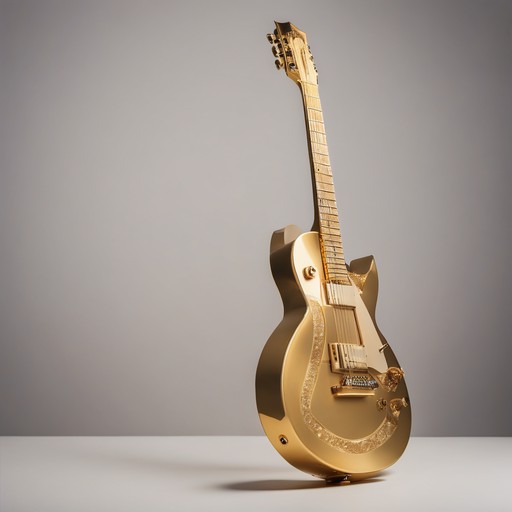} &
\includegraphics[width=0.22\linewidth]{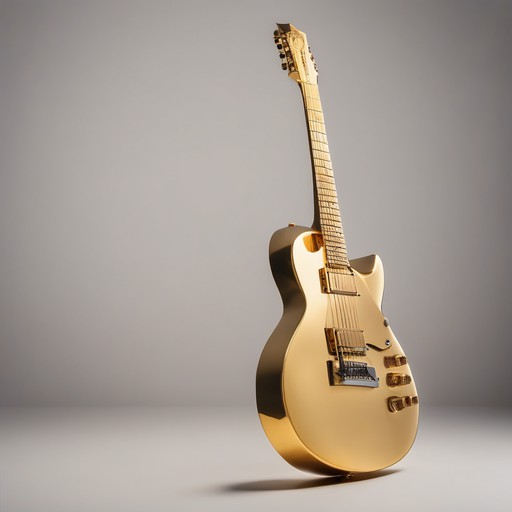} \\[2pt]

\raisebox{0.8cm}{\rotatebox{90}{\parbox{1.6cm}{\centering\scriptsize SDXL\\``blue apple''}}} &
\includegraphics[width=0.22\linewidth]{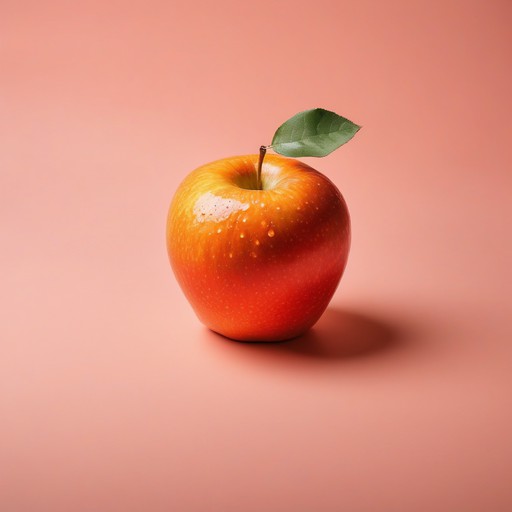} &
\includegraphics[width=0.22\linewidth]{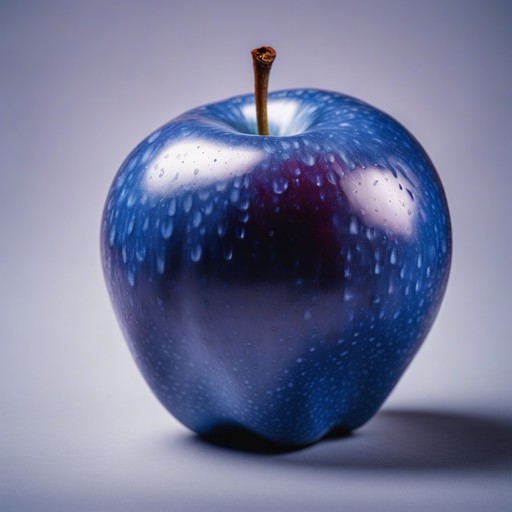} &
\includegraphics[width=0.22\linewidth]{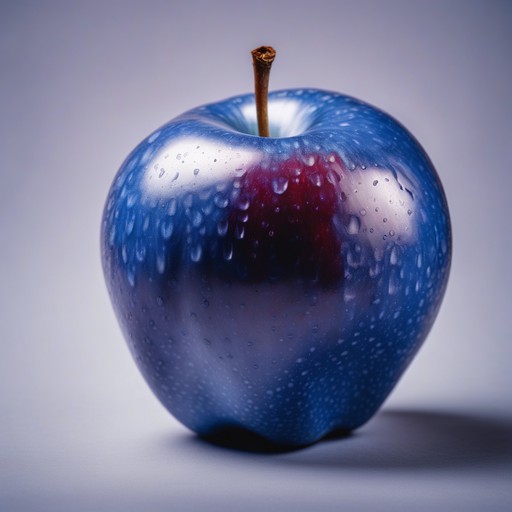} &
\includegraphics[width=0.22\linewidth]{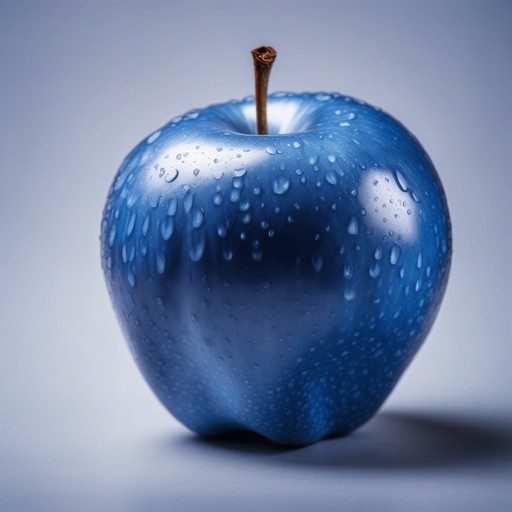} \\[2pt]

\raisebox{0.8cm}{\rotatebox{90}{\parbox{1.6cm}{\centering\scriptsize SD1.5\\``red horse''}}} &
\includegraphics[width=0.22\linewidth]{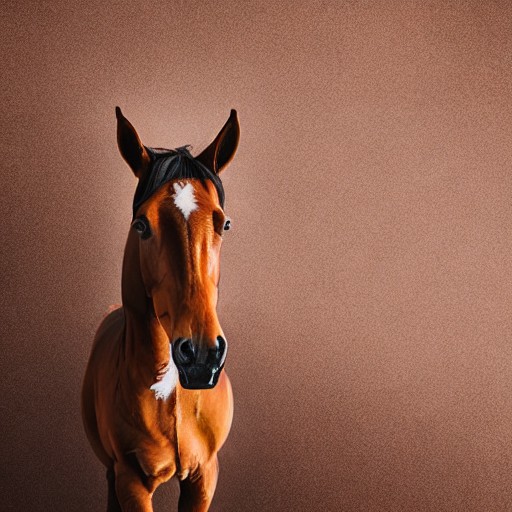} &
\includegraphics[width=0.22\linewidth]{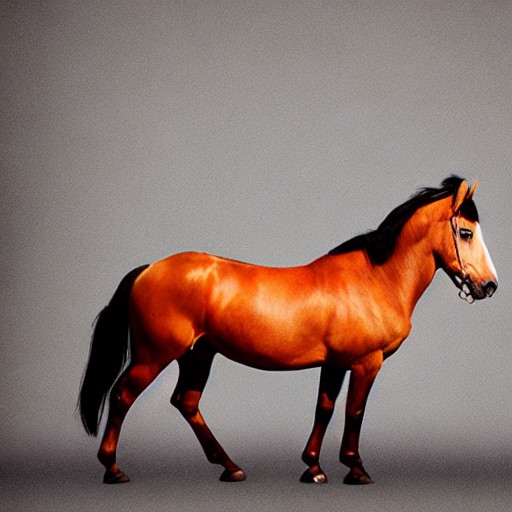} &
\includegraphics[width=0.22\linewidth]{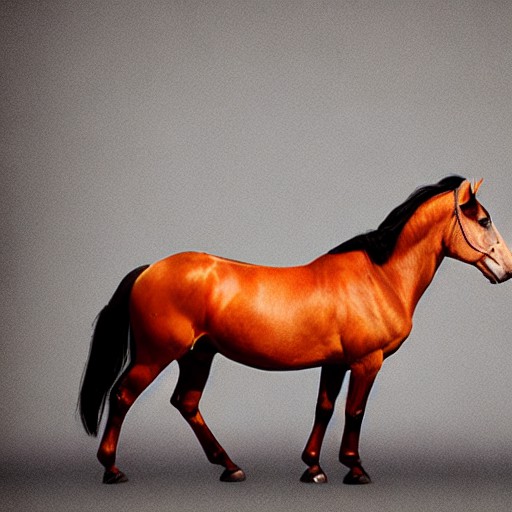} &
\includegraphics[width=0.22\linewidth]{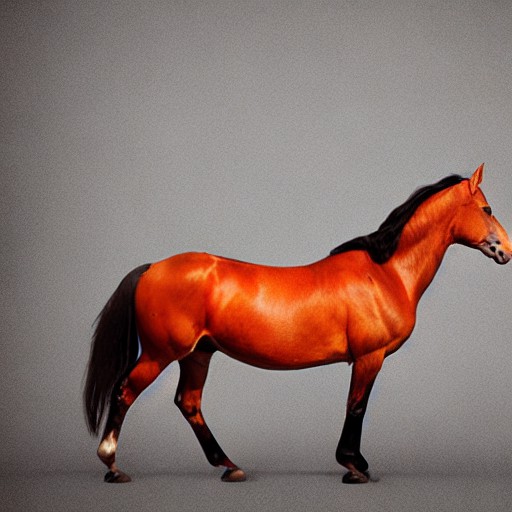} \\
\end{tabular}
\vspace{-0.3em}
\caption{Same-site (D2) vs.\ relocated (decoder) boost. D2 boost produces no visible change or degradation; decoder boost increases CHR in these examples. ``Gold guitar'': CHR 0.125$\to$0.114$\to$\textbf{0.244}. ``Blue apple'': 0.239$\to$0.224$\to$\textbf{0.372}. ``Red horse'': 0.148$\to$0.160$\to$\textbf{0.329}.}
\label{fig:qualitative}
\end{figure}

Selective boost at D2 has no detected CHR effect in either model (SDXL $p{=}0.4093$; SD1.5 $p{=}0.4819$). Decoder-minus-D2 is positive and nominally significant in SDXL ($+0.0074$, $p{=}0.0028$), but only directionally consistent in SD1.5 ($+0.0063$, $p{=}0.1882$). Decoder-minus-baseline is $+0.0061$ in SDXL (95\% CI $[0.0016,0.0111]$, $p{=}0.0168$, $q{=}0.0840$) and $+0.0084$ in SD1.5 ($[0.0008,0.0156]$, $p{=}0.0384$, $q{=}0.1344$). Neither survives within-model correction across blocks. Selective \textsc{attr-zero} shows no significant positive CHR effect; SD1.5 D0 has a small negative effect ($-0.0050$, $q{=}0.0014$). Failure to detect an effect at D2 is not an equivalence result. The separately configured timestep and fractional-attenuation sweeps are descriptive controls (Appendices~\ref{app:timestep} and~\ref{app:fractional}).

Broad ablation gives a different, architecture-dependent site profile (Figure~\ref{fig:fragility}); the joint boost--zeroing comparison is in Appendix~\ref{app:full_tables} (Figure~\ref{fig:polarity}). SDXL D2 has no detected change ($+0.0054$, $q{=}0.8273$), and SD1.5 D2 has a positive mean ($+0.0577$, $p{=}0.0356$) that does not survive FDR ($q{=}0.0623$). SDXL U0 shows a substantial decrease ($-0.1338$, $q{=}0.0010$), while SDXL D1 improves ($+0.0200$, $q{=}0.0330$). SD1.5 D0/D1 show substantial decreases ($q{=}0.0007$). Thus, the data support modality- and architecture-dependent responses, but do not establish that D2 ablation is beneficial or harmless in both models.

\begin{figure}[t]
\centering
\includegraphics[width=\linewidth]{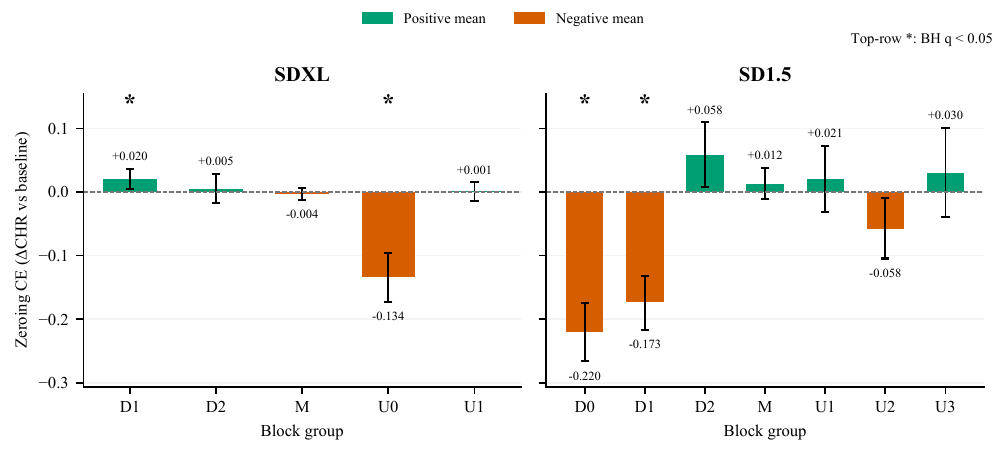}
\caption{Per-block CHR response to broad zeroing. Colors indicate the sign of the mean change; asterisks in the fixed top row indicate BH $q<0.05$ across active blocks within each model. Error bars show pointwise 95\% percentile-bootstrap confidence intervals (2,000 resamples). Each estimate uses 52 prompt-level differences averaged over three paired seeds. Corrected effects are positive at SDXL D1 and negative at SDXL U0 and SD1.5 D0/D1. See Table~\ref{tab:ce_zero_full}.}
\label{fig:fragility}
\end{figure}

\begin{table}[t]
\centering
\caption{CE under three modalities at D2 and decoder (both models). Bold: FDR $q{<}0.05$; ns denotes $q\geq0.05$. Boost $p$-values are uncorrected; no boost block survives within-model BH correction across active blocks (see text).}
\label{tab:main}
\small
\begin{tabular}{llccc}
\toprule
& & Selective additive & Selective subtractive & Broad subtractive \\
Model & Block & (boost) & (\texttt{attr\_zero}) & (zero) \\
\midrule
\multirow{2}{*}{SDXL} & D2 & $-0.0013$ (ns) & $-0.0011$ (ns) & $+0.0054$ (ns) \\
& U0 & $+0.0061$ ($p{=}0.0168$) & $-0.0002$ (ns) & $\mathbf{-0.1338}$ ($q{=}0.0010$) \\
\midrule
\multirow{2}{*}{SD1.5} & D2 & $+0.0021$ (ns) & $-0.0029$ (ns) & $+0.0577$ ($q{=}0.0623$) \\
& U1 & $+0.0084$ ($p{=}0.0384$) & $-0.0031$ (ns) & $+0.0207$ (ns) \\
\bottomrule
\end{tabular}
\end{table}

Together, the modality decomposition separates the largest positive signed DA mean at D2 from the largest positive boost CE mean at the decoder (Table~\ref{tab:main}). The decoder-minus-D2 contrast is supported in SDXL but not significant in SD1.5; decoder-minus-baseline does not survive per-block FDR in either model. Broad ablation has an architecture-dependent profile, including a large negative response at SDXL U0. These findings support evaluating diagnostic readouts and intervention operators separately, with the strength of each comparison stated explicitly.

\FloatBarrier
\section{Discussion}
\label{sec:discussion}

Taken together, the upstream and downstream analyses reveal a specific property of diffusion pipelines: as the conditioning representation flows from text encoder to denoiser cross-attention, the features that are most informative about \emph{which prompts will fail} and the features that are most responsive to \emph{interventions that correct failure} live in different places. Text-encoder residuals provide a diagnostic coordinate; D2 has the largest positive signed DA mean; and decoder blocks have the largest positive boost CE means. Together these measurements locate distinct diagnostic and intervention-response profiles along the pipeline. The site-level mismatch extends to the prompt level as well: CSI-stratified analysis shows no consistent positive correlation between pre-denoising stress severity and decoder boost response (SDXL partial $\rho{=}-0.17$, $p{=}0.007$; SD1.5 $\rho{=}+0.11$, ns), so higher diagnostic severity is not a consistently positive predictor of intervention response across models. Diagnosis and intervention are not aligned by default along the pipeline; whether they can be realigned depends on choices about objective (embedding repair, \S\ref{sec:results2}) and modality (per-block decomposition, \S\ref{sec:results3}), not merely site.

Two observations sharpen the dissociation. At D2, neither tested selective operator shows an FDR-significant CHR effect. In the descriptive per-timestep sweep, D2 has its largest DA mean and most negative boost CE mean during early denoising (Appendix~\ref{app:timestep}); these phase means do not establish a significant difference between phases. The small responses to selective interventions are compatible with either a distributed defect across D2's cross-attention output or a localized defect in a representation subspace that these operators do not effectively modify. At the decoder, U0's negative DA ($-0.0118$, FDR $q{=}0.00033$ in the SDXL boost baseline) indicates a reverse anchor--stress ABM contrast. This contrast measures an attention difference, not whether the decoder faithfully executes an upstream binding. Overriding existing binding with a competing signal remains one possible interpretation of the boost response, rather than a mechanism established by these comparisons.
These findings motivate evaluating diagnostic readouts and intervention operators jointly while selecting intervention sites using measured output responses.

Beyond our own protocol, the modality--site dissociation has implications for existing inference-time methods. Attend-and-Excite~\citep{chefer2023attend} and SynGen~\citep{rassin2024syngen} apply gradient-based corrections across all cross-attention layers simultaneously; our per-block decomposition suggests their effectiveness may concentrate at decoder blocks, because decoder blocks have the largest positive boost CE means in our sweeps. If the encoder-side contribution is zero or negative, all-layer methods may succeed \emph{despite} their encoder-side component rather than because of it---a testable prediction: restricting A\&E/SynGen updates to decoder-only cross-attention should preserve or improve performance while reducing compute, and restricting to encoder-only should degrade performance to baseline or worse.

More broadly, intervention design depends on how downstream computation uses a representation. Recent multimodal models address this relationship through answer-oriented latent visual prediction~\citep{li2026toward} and explicit coupling between semantic reasoning and 3D geometry~\citep{yu2025core3d,yu2026elsa3d}. Our results motivate evaluating such representation interfaces along two axes: how clearly they expose task-relevant difficulties and how effectively targeted edits change downstream behavior. Extending this analysis could clarify how representation structure and training objectives shape the choice of intervention site and operator.

\subsection{Scope and Interpretation} The central conclusion of this work is not that a particular intervention site universally solves compositional failures. Rather, our evidence establishes that diagnostic accessibility, representation modification, and semantic improvement should not be assumed to coincide. Accordingly, we do not claim (i) that all upstream interventions fail, (ii) that decoder-side intervention provides a universal repair mechanism, or (iii) that the observed diagnosis-control dissociation has been established for relational or role-binding compositions. Our results are supported for the evaluated attribute-object setting and the tested intervention modalities.

\section{Limitations}
\label{sec:limitations}

We distinguish established from suggestive findings. \textbf{Established}: (i)~CSI separates anchor from stress cross-model (ROC-AUC${=}1.0$, permutation null${=}0.49$, within-stress $\rho{=}0.68$), though partly reflecting the controlled protocol; (ii)~D2 has the largest positive signed DA mean in the six baseline datasets and differs from zero after FDR; these are not six independent replications; (iii)~selective subtraction shows no significant positive CHR effect across the reported blocks; (iv)~embedding repair reduces residuals without semantic improvement. \textbf{Suggestive}: (v)~decoder boost as a preferred selective site (baseline contrasts nominally significant in both models but not after per-block FDR; decoder-minus-D2 significant only in SDXL); (vi)~D2 defect structure (consistent with but not uniquely identified by our operators); (vii)~generality beyond chromatic binding.

Key caveats: boost CE is small ($+0.006$--$0.008$) and does not survive within-model FDR; CE is color-only (52 valid prompts per seed); SD3 is text-path only; two UNet architectures, no DiT; fixed intervention hyperparameters; CSI-based routing and a universal decoder repair interface unvalidated; broad-additive cell untested; WiCLP-aligned embedding repair not tested; object mask from baseline attention, held fixed across conditions. We characterize a modality-dependent structure, not a universal law. External-mask and background-color sensitivity checks further limit the interpretation to a small chromatic response rather than established object-local semantic repair (Appendix~\ref{app:color_sensitivity}).

\section{Conclusion}
\label{sec:conclusion}

We characterize compositional diagnosis and intervention through complementary measurements of text representations, cross-attention, and generated color agreement. Text-only CSI provides an upstream stress coordinate across SD1.5, SDXL, and the SD3 text path under the controlled anchor--stress protocol. Within the two tested UNets, the largest positive signed DA mean occurs at D2, while the largest positive selective-boost CE mean occurs at decoder blocks. A matched six-prompt localization study further links intervention location to output response, with downstream intervention increasing CHR as residual-minimizing adapters improve representation fit. Broad ablation reveals distinct architecture-dependent sensitivity profiles. Together, these findings provide an empirical site--modality map and motivate choosing intervention sites through output-response evaluation alongside diagnostic analysis. Extending this framework to broader compositional benchmarks, binding-aligned objectives, and DiT backbones offers concrete directions for future work. Code, CSI computation scripts, prompts, and seeds will be released on GitHub upon publication.

\clearpage
\section*{Acknowledgements}
Research reported in this publication was supported by DOE ASCR under Award Number DE-SC0022873, the National Institutes of Health under Award Number R01GM143789, and the Advanced Research Projects Agency for Health (ARPA-H) under Award Number D24AC00338-00. The content is solely the responsibility of the authors and does not necessarily represent the official views of the National Institutes of Health, the Department of Energy, or the Advanced Research Projects Agency for Health.

\bibliographystyle{plainnat}
\bibliography{merged_refs}

\clearpage
\appendix

\section*{Appendix Contents}
\begingroup
\small
\begin{description}
\setlength{\itemsep}{2pt}
\setlength{\parskip}{0pt}
\item[\ref{app:related_extended}] \hyperref[app:related_extended]{Extended Related Work} \hfill \pageref{app:related_extended}
\item[\ref{app:csi}] \hyperref[app:csi]{CSI Definition and Discovery Details} \hfill \pageref{app:csi}
\item[\ref{app:embed}] \hyperref[app:embed]{Embedding-Level Repair Details} \hfill \pageref{app:embed}
\item[\ref{app:modality}] \hyperref[app:modality]{Intervention Modality Matrix} \hfill \pageref{app:modality}
\item[\ref{app:da_stability}] \hyperref[app:da_stability]{DA Stability Across Conditions} \hfill \pageref{app:da_stability}
\item[\ref{app:samesite}] \hyperref[app:samesite]{Same-Site vs Relocated Comparison} \hfill \pageref{app:samesite}
\item[\ref{app:full_tables}] \hyperref[app:full_tables]{Full Per-Block CE Tables} \hfill \pageref{app:full_tables}
\item[\ref{app:ce_boost}] \hyperref[app:ce_boost]{CE under selective additive boost} \hfill \pageref{app:ce_boost}
\item[\ref{app:ce_attr_zero}] \hyperref[app:ce_attr_zero]{CE under selective subtractive attr\_zero} \hfill \pageref{app:ce_attr_zero}
\item[\ref{app:ce_broad_zero}] \hyperref[app:ce_broad_zero]{CE under broad subtractive zeroing} \hfill \pageref{app:ce_broad_zero}
\item[\ref{app:abm}] \hyperref[app:abm]{ABM Metric and Validation} \hfill \pageref{app:abm}
\item[\ref{app:chr}] \hyperref[app:chr]{Color Hit Rate Details} \hfill \pageref{app:chr}
\item[\ref{app:color_sensitivity}] \hyperref[app:color_sensitivity]{External-Mask Sensitivity Checks} \hfill \pageref{app:color_sensitivity}
\item[\ref{app:blocks}] \hyperref[app:blocks]{Block Inventories} \hfill \pageref{app:blocks}
\item[\ref{app:timestep}] \hyperref[app:timestep]{Per-Timestep Phase Analysis} \hfill \pageref{app:timestep}
\item[\ref{app:power}] \hyperref[app:power]{Power Analysis for Decoder Boost} \hfill \pageref{app:power}
\item[\ref{app:prompts}] \hyperref[app:prompts]{Prompt Examples} \hfill \pageref{app:prompts}
\item[\ref{app:fractional}] \hyperref[app:fractional]{Fractional Attenuation Curve} \hfill \pageref{app:fractional}
\item[\ref{app:compute}] \hyperref[app:compute]{Compute Resources} \hfill \pageref{app:compute}
\end{description}
\endgroup
\clearpage

\section{Extended Related Work}
\label{app:related_extended}

A growing body of work examines compositional generation and evaluation in controlled settings. \citet{okawa2023compositional} show that compositional abilities in diffusion models emerge multiplicatively rather than additively, consistent with our finding that compositional entanglement in text representations resists simple additive decomposition. \citet{marioriyad2024diffusion} provide a systematic evaluation comparing diffusion and autoregressive architectures on compositionality, reinforcing that compositional failures are architecture-general. Broader evaluation suites---T2I-CompBench++~\citep{huang2023t2icompbench}, Winoground~\citep{thrush2022winoground}, CREPE~\citep{ma2023crepe}, and SugarCrepe~\citep{hsieh2024sugarcrepe}---measure compositional success at the output level across attribute, relation, and object categories; our work is complementary, diagnosing \emph{where} inside the pipeline these failures originate rather than \emph{whether} they occur.

Object-centric and disentangled representation learning~\citep{locatello2020slot} provides a conceptual parallel: slot-based models enforce compositional structure by design, whereas text-to-image diffusion models inherit whatever compositional structure CLIP-family encoders provide. This emphasis on representation structure also appears in multimodal reasoning: Cognitive Supersensing learns latent visual predictions aligned with answer-image embeddings~\citep{li2026toward}, while CoRe3D organizes geometric reasoning over spatially localized 3D latents~\citep{yu2025core3d}. ELSA3D uses anchors to connect semantic cues to geometric evidence at selected scales~\citep{yu2026elsa3d}. Energy-based compositional generation~\citep{du2023reduce} offers an alternative at the sampling stage through explicit concept combination. Our per-block analysis complements these perspectives by evaluating how diagnostic readouts and intervention responses vary within existing diffusion pipelines.

\section{CSI Definition and Discovery Details}
\label{app:csi}

The non-factorization residual $R$ (Eq.~1) is computed under three low-complexity composition models: additive, low-rank linear, and low-rank bilinear. Composition parameters are fit within training folds only (leave-one-group-out); test groups are never used for fitting.

\paragraph{Encoder extraction.}
SD1.5 uses the sequence mean of the final CLIP hidden states (768 dimensions). SDXL concatenates the sequence means of the final CLIP-L (768) and OpenCLIP-G (1280) hidden states with the projected pooled OpenCLIP-G output (1280), yielding one 3328-dimensional vector. A single composition model is fit to this concatenated representation; encoder-wise CSI scores are not averaged. SD3 uses only the sequence mean of the final T5-XXL hidden states, not joint CLIP/T5 conditioning. Each final vector is $\ell_2$-normalized.

The three composition models are:
\begin{itemize}
\item \textbf{Additive}: $\hat{h} = h(p_\text{attr}) + h(p_\text{obj}) + h(p_\text{ctx})$.
\item \textbf{Low-rank linear}: $\hat{h} = W_\text{attr} h(p_\text{attr}) + W_\text{obj} h(p_\text{obj}) + W_\text{ctx} h(p_\text{ctx})$, where $W_i$ are rank-$r$ matrices fit per LOGO fold.
\item \textbf{Low-rank bilinear}: adds a rank-$r$ bilinear interaction term between attribute and object representations.
\end{itemize}

For span geometry, define component matrix $B=[h(p_\text{attr}), h(p_\text{obj}), h(p_\text{ctx})] \in \mathbb{R}^{d\times 3}$ and $f=h(p_\text{full})$. The projected vector is $\hat{f} = B(B^\top B + \lambda I)^{-1}B^\top f$, yielding:
\begin{equation}
R_\text{span} = \frac{\lVert f - \hat{f} \rVert_2}{\lVert f \rVert_2 + \varepsilon}, \qquad
\theta_\text{span} = \arccos\!\left(\frac{\lVert \hat{f} \rVert_2}{\lVert f \rVert_2 + \varepsilon}\right).
\end{equation}

CSI aggregates text-residual and span-geometry features:
\begin{equation}
\mathrm{CSI}_\text{text} = \tfrac{1}{2}(z(r_\text{lin}) + z(r_\text{bilin})), \quad
\mathrm{CSI}_\text{span} = \tfrac{1}{2}(z(R_\text{span}) + z(\theta_\text{span})),
\end{equation}
\begin{equation}
\mathrm{CSI} = \tfrac{1}{2}(\mathrm{CSI}_\text{text} + \mathrm{CSI}_\text{span}),
\end{equation}
where $z(\cdot)$ is robust standardization using anchor-only statistics per model. No weights are tuned on downstream labels.
Specifically, $z(v)=(v-\operatorname{median}_{A}v)/(s_A+10^{-8})$, where $s_A$ is the unscaled median absolute deviation on anchors; a non-finite or near-zero value ($<10^{-8}$) falls back to the population standard deviation, and then to $1$ if still degenerate.

\paragraph{Calibration and diagnostic scope.}
The discovery scores use all anchors within each model for this standardization, separately from fold-wise composition fitting. They therefore do not establish fully held-out calibration or deployment performance. Pair-level partitions can also share an anchor text across groups and do not establish generalization to unseen anchor texts. The span features include internal mid-attention quantities, so only $\mathrm{CSI}_{\mathrm{text}}$ supports a strictly pre-denoising interpretation; combined-CSI associations are retrospective diagnostics and should not be attributed to a text-only score without recomputation.

Discovery is assessed by a Ladder Consistency Criterion (LCC): monotonic decile trend, first-to-last delta CI excluding 0, and consistent direction across models. The stricter Correlation Threshold Sensitivity check ($\rho \geq 0.90$) is reported as sensitivity analysis only.

Several robustness analyses confirm that the CSI signal is not an artifact. Group-preserving label-permutation yields null ROC-AUC $= 0.49 \pm 0.01$ against a true ROC-AUC of $0.89$--$1.00$ (all empirical $p = 0.001$), and CSI ROC-AUC${=}1.0$ holds in all four conflict-type strata (chromatic oddness, material--object, geometric pattern, state--environment), so the signal is not driven by a single attribute class.

We further apply a three-level covariate control protocol. Level~1 uses covariate-adjusted analyses controlling for perplexity, subtoken count, attribute frequency, prompt length, and template family. Level~2 uses near-matched subsets ($n_\text{pairs}{=}140$, canonical tier); near-matched LOGO residual effects remain significant (linear mean diff${=}0.543$, $q{<}0.001$; bilinear mean diff${=}0.681$, $q{<}0.001$), and strict ($n{=}112$) and relaxed ($n{=}178$) tiers preserve the same pattern. Level~3 bounds interpretation: these controls weaken lexical/tokenization explanations but do not eliminate all language-side priors.

A predictive-asymmetry breakdown clarifies which CSI families carry the signal. Within-dataset family ablation across 7 partitions yields combined ROC-AUC${=}0.9999$, text-only${=}0.9999$, and span-only${=}0.6695$; cross-model transfer across 30 directions yields text-only mean ROC-AUC${=}0.9999$, combined${=}0.9986$, and span-only${=}0.5745$. Text families dominate; span-only is near chance in transfer.

\section{Embedding-Level Repair Details}
\label{app:embed}

Two low-rank adapter families (rank 32, AdamW, 800 epochs, MSE objective) are evaluated. The pooled-adapter representation experiment has 20 representation-evaluable prompts ($\Delta_\text{repr}{=}-0.0998$, $q{=}0.0024$); the token-adapter representation experiment has 19 ($-0.1836$, $q{=}0.0008$). Each has seven CHR-evaluable prompts with observed zero paired change ($p{=}q{=}1$); this small-sample observation does not establish equivalence. Within each run, BH correction covers representation residual and CHR. Separate full-scale rank-sweep runs evaluate 216 representation prompts and 52 CHR prompts: pooled adapters show no significant positive CHR effect, while token adapters have negative CHR effects. Their residual improvements are not monotonic in rank; these rank-sweep experiments use different prompt sets from the smaller representation experiments. No correction across all rank configurations is claimed. WiCLP~\citep{zarei2024wiclp} uses a binding-aligned objective, so our findings remain specific to the tested residual-minimizing objectives.

For the six-prompt localization study, CHR is evaluated on six prompts after averaging five seeds; downstream ABM uses 24 prompts. Downstream BH correction covers early ABM, mid ABM, and CHR, giving CHR $q{=}0.046875$. The embedding arm uses its separate two-endpoint family (residual and CHR), giving CHR $q{=}0.03125$. With six prompts, exact two-sided sign-flip inference enumerates 64 sign patterns. An exploratory paired comparison of CHR gains is $+0.0952$ (95\% CI $[0.0434,0.1509]$, raw $p{=}0.03125$); it was not a prespecified superiority test. Identical baseline CHR values do not by themselves verify identical images or masks.

\section{Intervention Modality Matrix}
\label{app:modality}

Table~\ref{tab:modality_mat} summarizes the selectivity$\times$sign intervention matrix. We test five intervention families across two pipeline stages (embedding and denoiser). At the denoiser level, three of four cells are tested; the ``broad additive'' cell (uniformly scaling all post-softmax attention entries) is a mathematical identity under renormalization and is therefore untested. A proper broad-additive control would require pre-softmax logit manipulation, which our hook architecture does not support.

\paragraph{Denoiser intervention settings.}
For the full per-block SD1.5 and SDXL experiments, selective boost uses intervention strength $4.0$. The object-attention mask used for selective interventions is thresholded at quantile $0.80$; the object mask used for CHR scoring also uses quantile $0.80$ (Appendix~\ref{app:chr}). Generation uses 40 denoising timesteps with classifier-free guidance (CFG) scale $7.5$. Interventions are active throughout all 40 timesteps in the full per-block runs, with these settings held fixed across block comparisons. The separately reported phase-restricted analysis changes the intervention window as described in Appendix~\ref{app:timestep}. Boost and selective \texttt{attr\_zero} use five paired seeds per prompt, while broad zeroing uses three. The strength parameter applies to boost; selective and broad zeroing use the removal operators specified above. Statistical testing is described in \S\ref{sec:setup}.

\begin{table}[h]
\centering
\small
\caption{Selectivity $\times$ sign intervention matrix. Three denoiser cells and two embedding cells are tested; broad additive is a no-op under post-softmax renormalization.}
\label{tab:modality_mat}
\begin{tabular}{@{}l|c|c@{}}
& \textbf{Selective} (token-level) & \textbf{Broad} (block/pool-wide) \\
\hline
\textbf{Embedding additive} & Token adapter (MSE) & Pooled adapter \\
\hline
\textbf{Denoiser additive} & Boost (attr $\times$ obj mask) & Untested (identity) \\
\hline
\textbf{Denoiser subtractive} & Attr\_zero (attr $\times$ obj mask) & Zero (uniform $1/k$) \\
\end{tabular}
\end{table}

\section{DA Stability Across Conditions}
\label{app:da_stability}

D2 has the largest positive signed DA mean in all six baseline datasets, and its contrast against zero survives within-model, within-run FDR (Table~\ref{tab:da_stab}). Boost and attr-zero baseline statistics coincide and must not be treated as independent replications. No prompt IDs span multiple group IDs in these files, but shared anchor texts or base identities remain a separate dependence question.

\begin{table}[h]
\centering
\caption{DA at D2 across all experimental conditions. D2 has the largest positive signed DA mean in each condition; the reported tests compare D2 with zero, not with other blocks.}
\label{tab:da_stab}
\small
\begin{tabular}{llcc}
\toprule
Model & Condition & DA(D2) & FDR $q$ \\
\midrule
SDXL & Boost baseline & $+0.0039$ & $0.0174$ \\
SDXL & Zero baseline & $+0.0035$ & $0.0332$ \\
SDXL & Attr\_zero baseline & $+0.0039$ & $0.0174$ \\
SD1.5 & Boost baseline & $+0.0282$ & $0.00028$ \\
SD1.5 & Zero baseline & $+0.0296$ & $0.00028$ \\
SD1.5 & Attr\_zero baseline & $+0.0282$ & $0.00028$ \\
\bottomrule
\end{tabular}
\end{table}

\section{Same-Site vs Relocated Comparison}
\label{app:samesite}

Table~\ref{tab:ss_reloc} reports paired comparisons from the full per-block boost runs after averaging five seeds within each of 52 prompts. Decoder-minus-D2 is nominally significant in SDXL ($p{=}0.0028$), but not SD1.5 ($p{=}0.1882$). Neither D2 baseline contrast is significant; this does not establish equivalence.

\begin{table}[h]
\centering
\caption{Same-site (D2) vs relocated (decoder) under boost. Tests use 52 prompt means after pairing five seeds; displayed significance is nominal, before multiplicity correction.}
\label{tab:ss_reloc}
\small
\begin{tabular}{llccc}
\toprule
Model & Comparison & $\Delta$ CHR & $p$ & Sig.? \\
\midrule
\multirow{3}{*}{SDXL}
& Same-site (D2) vs baseline & $-0.0013$ & $0.4093$ & No \\
& Relocated (U0) vs baseline & $+0.0061$ & $0.0168$ & Yes \\
& \textbf{Relocated vs same-site} & $\mathbf{+0.0074}$ & $\mathbf{0.0028}$ & \textbf{Yes} \\
\midrule
\multirow{3}{*}{SD1.5}
& Same-site (D2) vs baseline & $+0.0021$ & $0.4819$ & No \\
& Relocated (U1) vs baseline & $+0.0084$ & $0.0384$ & Yes \\
& Relocated vs same-site & $+0.0063$ & $0.1882$ & No \\
\bottomrule
\end{tabular}
\end{table}

\section{Full Per-Block CE Tables}
\label{app:full_tables}

Tables~\ref{tab:ce_boost_full}, \ref{tab:ce_atz_full}, and~\ref{tab:ce_zero_full} report prompt-level CE at all active blocks. Boost estimates use the full per-block runs with 52 CHR-evaluable prompts. Boost and broad-zero tables display raw $p$, with bold effects only when within-run BH $q<0.05$; attr-zero displays $q$. No boost block survives FDR. Attr-zero has no significant positive effect, with a negative effect at SD1.5 D0. Broad zeroing improves SDXL D1 and decreases SDXL U0 and SD1.5 D0/D1 after FDR; SD1.5 D2 and U2 do not survive correction.

\begin{figure}[!htbp]
\centering
\includegraphics[width=\linewidth]{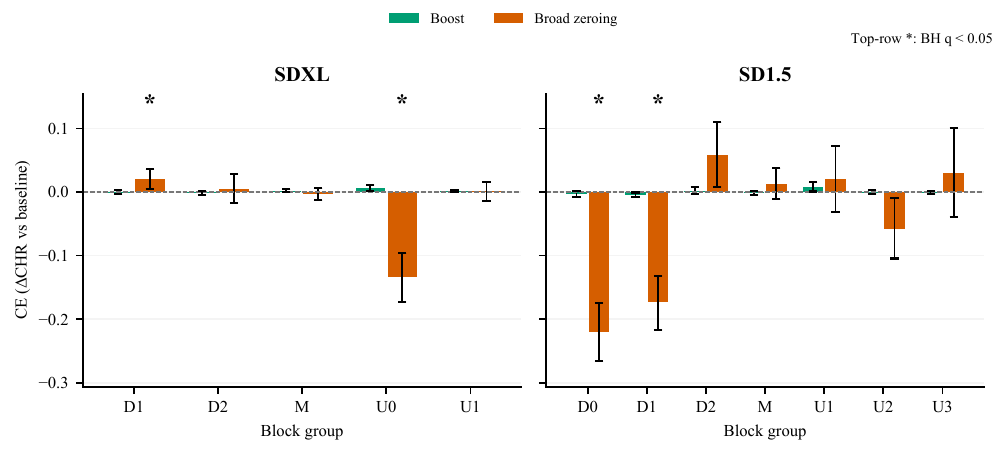}
\caption{Per-block CE under selective boost (green) and broad zeroing (orange), relative to the corresponding baseline. Error bars show pointwise 95\% percentile-bootstrap confidence intervals (2,000 resamples). Estimates use 52 prompt means with five paired seeds for boost and three for zeroing. Asterisks in the fixed top row indicate BH $q<0.05$ across active blocks within each model and intervention run; they test each effect against zero, not the difference between modalities. No boost block survives correction. The shared vertical scale emphasizes the larger zeroing effects; boost is shown in detail in Figures~\ref{fig:da_ce} and~\ref{fig:samesite}. See Tables~\ref{tab:ce_boost_full} and~\ref{tab:ce_zero_full}.}
\label{fig:polarity}
\end{figure}

Figure~\ref{fig:polarity} compares the boost and broad-zeroing profiles on a shared scale.

\subsection{CE under selective additive boost (432 prompts $\times$ 5 seeds)}
\label{app:ce_boost}

Table~\ref{tab:ce_boost_full} reports per-block CE under selective additive boost for SDXL and SD1.5. The largest positive mean responses occur at decoder blocks U0 and U1, respectively, but no block survives FDR correction.

\begin{table}[h]
\centering
\caption{CE$_\text{boost}$ per block. Columns show raw $p$-values. No block survives FDR correction; the decoder peaks (SDXL U0, SD1.5 U1) are nominally significant versus baseline, with consistent cross-model direction.}
\label{tab:ce_boost_full}
\small
\begin{tabular}{lcccc}
\toprule
Block & SDXL CE & SDXL $p$ & SD1.5 CE & SD1.5 $p$ \\
\midrule
D0 & -- & -- & $-0.0031$ & 0.1854 \\
D1 & $-0.0002$ & 0.8826 & $-0.0041$ & 0.0182 \\
D2 & $-0.0013$ & 0.4093 & $+0.0021$ & 0.4819 \\
M & $+0.0018$ & 0.1242 & $-0.0015$ & 0.2763 \\
U0 & $+0.0061$ & 0.0168 & -- & -- \\
U1 & $+0.0011$ & 0.1788 & $+0.0084$ & 0.0384 \\
U2 & -- & -- & $+0.0002$ & 0.8658 \\
U3 & -- & -- & $-0.0005$ & 0.6401 \\
\bottomrule
\end{tabular}
\end{table}

\subsection{CE under selective subtractive attr\_zero (432 $\times$ 5 seeds, both models full-scale)}
\label{app:ce_attr_zero}

Table~\ref{tab:ce_atz_full} reports per-block CE under selective attribute-token zeroing for both models, together with BH-adjusted $q$-values. No block shows a significant positive effect; SD1.5 D0 shows a significant negative effect.

\begin{table}[h]
\centering
\caption{CE$_\text{attr\_zero}$ per block with prompt-level BH $q$. SD1.5 D0 has a small significant negative effect; no block has a significant positive effect.}
\label{tab:ce_atz_full}
\small
\begin{tabular}{lcccc}
\toprule
Block & SDXL CE & SDXL FDR $q$ & SD1.5 CE & SD1.5 FDR $q$ \\
\midrule
D0 & -- & -- & $\mathbf{-0.0050}$ & 0.0014 \\
D1 & $+0.0003$ & 0.8514 & $-0.0010$ & 0.4669 \\
D2 & $-0.0011$ & 0.8514 & $-0.0029$ & 0.0602 \\
M & $-0.0003$ & 0.8514 & $-0.0022$ & 0.1225 \\
U0 & $-0.0002$ & 0.8514 & -- & -- \\
U1 & $+0.0007$ & 0.8514 & $-0.0031$ & 0.0970 \\
U2 & -- & -- & $-0.0018$ & 0.2634 \\
U3 & -- & -- & $-0.0031$ & 0.2312 \\
\bottomrule
\end{tabular}
\end{table}

\subsection{CE under broad subtractive zeroing (432 $\times$ 3 seeds)}
\label{app:ce_broad_zero}

Table~\ref{tab:ce_zero_full} reports per-block CE under broad subtractive zeroing for both models. After FDR correction, zeroing increases CHR at SDXL D1 and decreases CHR at SDXL U0 and SD1.5 D0/D1.

\begin{table}[h]
\centering
\caption{CE$_\text{zero}$ per block. Bold: FDR $q < 0.05$. Columns show raw $p$-values; boldface indicates FDR-corrected significance.}
\label{tab:ce_zero_full}
\small
\begin{tabular}{lcccc}
\toprule
Block & SDXL CE & SDXL $p$ & SD1.5 CE & SD1.5 $p$ \\
\midrule
D0 & -- & -- & $\mathbf{-0.2199}$ & 0.0002 \\
D1 & $\mathbf{+0.0200}$ & 0.0132 & $\mathbf{-0.1731}$ & 0.0002 \\
D2 & $+0.0054$ & 0.6619 & $+0.0577$ & 0.0356 \\
M & $-0.0037$ & 0.4395 & $+0.0123$ & 0.3511 \\
U0 & $\mathbf{-0.1338}$ & 0.0002 & -- & -- \\
U1 & $+0.0008$ & 0.9126 & $+0.0207$ & 0.4597 \\
U2 & -- & -- & $-0.0576$ & 0.0286 \\
U3 & -- & -- & $+0.0296$ & 0.4325 \\
\bottomrule
\end{tabular}
\end{table}

\section{ABM Metric and Validation}
\label{app:abm}

ABM is computed per cross-attention head per timestep. For probability matrix $P \in \mathbb{R}^{q \times k}$:
(1)~object map $\mathbf{o} = \sum_{j \in \text{obj}} P_{:,j}$;
(2)~attribute map $\mathbf{a} = \sum_{j \in \text{attr}} P_{:,j}$;
(3)~mask $M_i = \mathbf{1}[\mathbf{o}_i \geq Q_{0.80}(\mathbf{o})]$;
(4)~ABM $= \sum_i \mathbf{a}_i M_i / \sum_i \mathbf{a}_i$.
Averaged across heads, timesteps, and seeds.

To validate ABM as a proxy for compositional binding, we correlate it with the CHR endpoint. Stress-prompt ABM at D2 correlates negatively with stress CHR: SDXL $\rho{=}-0.235$ ($p{=}0.0001$); SD1.5 $\rho{=}-0.124$ ($p{=}0.046$). The correlation is modest ($R^2 \approx 5\%$), indicating ABM is a meaningful but partial proxy. DA (anchor$-$stress ABM) is weakly positively correlated with stress CHR ($\rho{=}+0.17$); this is an arithmetic consequence: anchor ABM is uncorrelated with stress CHR ($\rho \approx 0$), so DA inherits its structure from the stress ABM term.

\section{Color Hit Rate Details}
\label{app:chr}

The object-local color hit rate (CHR) measures target-color agreement within an attention-derived object region. In the full per-block runs, 52 of 216 stress prompts have finite CHR across the required seeds; the other 164 have no target color recorded by the scorer. Baseline diagnostics did not attribute these exclusions to invalid masks. Absence of a recognized target color is not a verified semantic category label. Seed-level scores are paired within prompt and then averaged, giving 52 statistical units for CE, rather than $52\times\text{seeds}$. The separate six-prompt localization study is reported in Appendix~\ref{app:embed}; it is not pooled with these per-block results.

11 color prototypes (RGB, 0--1 scale): red (0.87,0.21,0.21), orange (0.95,0.56,0.20), yellow (0.93,0.86,0.22), green (0.23,0.70,0.31), blue (0.22,0.40,0.90), purple (0.58,0.35,0.86), pink (0.95,0.60,0.76), brown (0.55,0.36,0.24), black (0.08,0.08,0.08), white (0.95,0.95,0.95), gray (0.55,0.55,0.55). Distance threshold: 0.35. Object mask quantile: 0.80. Minimum mask fraction: 0.002. Scoring eligibility follows the recorded target colors and finite endpoint values; a separate prompt taxonomy must not be inferred from missing CHR alone.

\section{External-Mask Sensitivity Checks}
\label{app:color_sensitivity}

A separate held-out SDXL evaluation used 220 prompts, five seeds, and 20 active cross-attention modules per site (3.01\% mean-JS mismatch). U0--baseline CHR was $+0.003775$ (95\% CI $[+0.002328,+0.005240]$), and U0--D2 was $+0.008149$ ($[+0.006190,+0.010142]$). These estimates use prompt-level inference and are separate from the full per-block experiments. With baseline-only GroundingDINO+SAM masks shared across all arms, the U0--baseline interval crossed zero, while U0--D2 remained positive. A continuous target-color score (negative mean CIEDE2000) showed comparable U0--baseline changes inside the mask ($+0.154668$) and in the background ($+0.156614$); the inside-minus-background contrast was $-0.001946$ (95\% CI $[-0.055536,+0.054245]$). Thus increased CHR alone does not establish localized semantic repair.

\section{Block Inventories}
\label{app:blocks}

SDXL and SD1.5 have different numbers of cross-attention modules per block group (Table~\ref{tab:block_inv}). Counting cross-attention (\texttt{attn2}) only, SDXL concentrates modules in D2 (20) and U0 (30); the SD1.5 module inventory counts all patched attention modules (2--6 per group), including self-attention, rather than verified cross-attention-only modules. D0 has no cross-attention in SDXL; U0 has none in SD1.5. Post-hoc module normalization alone does not rule out intervention-exposure differences between sites.

\begin{table}[h]
\centering
\caption{Module inventories per block group. SDXL counts cross-attention only; SD1.5 counts all patched attention modules. The two columns use different counting conventions and must not be compared as intervention exposure.}
\label{tab:block_inv}
\small
\begin{tabular}{lcc}
\toprule
Block & SDXL & SD1.5 \\
\midrule
D0 & 0 & 4 \\
D1 & 4 & 4 \\
D2 & 20 & 4 \\
M & 10 & 2 \\
U0 & 30 & 0 \\
U1 & 6 & 6 \\
U2 & 0 & 6 \\
U3 & -- & 6 \\
\bottomrule
\end{tabular}
\end{table}

\section{Per-Timestep Phase Analysis}
\label{app:timestep}

To examine whether all-timestep averaging masks a phase-specific response at D2, we run boost restricted to early ($\leq 33\%$), mid ($33$--$66\%$), or late ($>66\%$) denoising phases separately (Table~\ref{tab:timestep}). D2's DA is strongest in the early phase ($+0.016$), yet boost CE is most negative there ($-0.010$). The mid-phase D2 mean is slightly positive ($+0.001$); this descriptive table does not establish a significant benefit. U0 boost CE is also strongest in the early phase ($+0.008$), consistent with compositional decisions being made early.

\begin{table}[h]
\centering
\caption{Per-phase DA and CE at D2 (SDXL, 100 prompts $\times$ 3 seeds).}
\label{tab:timestep}
\small
\begin{tabular}{lccc}
\toprule
Phase & D2 DA & D2 CE (boost) & U0 CE (boost) \\
\midrule
Early & $+0.016$ & $-0.010$ & $+0.008$ \\
Mid & $+0.013$ & $+0.001$ & $+0.002$ \\
Late & $+0.008$ & $-0.000$ & $+0.001$ \\
\bottomrule
\end{tabular}
\end{table}

\section{Power Analysis for Decoder Boost}
\label{app:power}

We address whether the FDR non-significance of the decoder-vs-baseline boost reflects effect absence or sample-size limitation. For each model we compute the per-prompt effect size $d_z$ (mean paired difference divided by its standard deviation) on the color-stress subset and the power of a paired test at the current sample size.

Effect size on the full color-stress subset ($n{=}52$ stress prompts $\times$ 5 seeds, aggregated to per-prompt means) is $d_z{=}0.345$ for SDXL (U0) and $d_z{=}0.297$ for SD1.5 (U1). These are conventionally classified as small-to-moderate. Power at the current sample size is $0.685$ (SDXL) and $0.557$ (SD1.5). Reaching $80\%$ power would require $68$ and $91$ color stress prompts respectively---i.e., $16$ to $39$ additional prompts beyond our current $52$. BH thresholds depend on the ranks of all tested p-values and do not impose a fixed factor-of-$m$ penalty on every comparison. These descriptive power estimates do not establish why a result fails FDR.

The six-prompt localization study has prompt-level $d_z{=}1.04$ for downstream CHR. This selected subset uses a separate protocol; its effect size does not establish that higher CSI predicts greater response or that full-set averaging merely dilutes a common effect.

These power calculations are conditional on the observed effect sizes and do not establish that the underlying effect is nonzero or that limited power explains FDR non-significance. The paired same-site versus relocated comparison is reported separately from the per-block comparisons.

\begin{table}[h]
\centering
\caption{Effect size and power for decoder boost at the full color-stress subset.}
\label{tab:power}
\small
\begin{tabular}{lccccc}
\toprule
Model & Block & $d_z$ (per-prompt) & $n_{\text{prompt}}$ & Power & $n$ for 80\% power \\
\midrule
SDXL  & U0 & $0.345$ & $52$ & $0.685$ & $68$ \\
SD1.5 & U1 & $0.297$ & $52$ & $0.557$ & $91$ \\
\bottomrule
\end{tabular}
\end{table}

\section{Prompt Examples}
\label{app:prompts}

Table~\ref{tab:prompt_ex} shows representative anchor--stress pairs from the 432-prompt dataset. Each group shares the same object and context; only the target attribute changes. The stress attribute is selected from a lower-frequency pool to induce compositional rarity. Conflict types span four categories: chromatic oddness (e.g., blue pear), material--object (glass cat, metallic banana), geometric pattern (striped refrigerator), and state--environment (dry horse).

\begin{table}[h]
\centering
\caption{Representative anchor--stress pairs.}
\label{tab:prompt_ex}
\small
\begin{tabular}{lll}
\toprule
Group & Anchor & Stress \\
\midrule
B001 & a studio photo of a black cat & a studio photo of a glass cat \\
B005 & a studio photo of a yellow banana & a studio photo of a metallic banana \\
B028 & a studio photo of a green pear & a studio photo of a blue pear \\
B045 & a studio photo of a brown guitar & a studio photo of a gold guitar \\
\bottomrule
\end{tabular}
\end{table}

\section{Fractional Attenuation Curve}
\label{app:fractional}

To examine whether partial attenuation behaves differently from full attribute zeroing ($\alpha{=}0$), we test fractional attenuation at $\alpha \in \{0.00, 0.25, 0.50, 0.75\}$ on SDXL (432 prompts $\times$ 3 seeds), where $\alpha$ scales the attribute-token attention weight and $\alpha{=}1.0$ is the unmodified baseline. Table~\ref{tab:fractional} reports CE at D2 and U0 (decoder) at each level.

All reported mean CHR changes have magnitude below $0.002$. D2's mean increases across the four attenuation levels, whereas U0 shows small, non-monotonic changes around zero. These descriptive means are smaller than the SDXL U0 boost mean from the full per-block runs ($+0.0061$, raw $p{=}0.0168$, BH $q{=}0.0840$). For a within-model broad-zeroing reference, SDXL D2 has mean CE $+0.0054$ ($q{=}0.8273$); the larger SD1.5 D2 mean ($+0.0577$, $q{=}0.0623$) comes from a different architecture and should not be treated as a matched comparator. The attenuation table does not report prompt-level confidence intervals or corrected tests, so it does not establish equivalence to zero or confirm the competing-signal interpretation discussed in \S\ref{sec:discussion}.

\begin{table}[h]
\centering
\caption{Fractional attenuation mean CE at D2 and U0 (SDXL, 156 paired prompt--seed observations per level). These are descriptive means; prompt--seed observations are not independent prompts.}
\label{tab:fractional}
\small
\begin{tabular}{ccc}
\toprule
$\alpha$ & D2 CE & U0 CE \\
\midrule
0.00 & $-0.0006$ & $+0.0002$ \\
0.25 & $+0.0003$ & $-0.0006$ \\
0.50 & $+0.0008$ & $+0.0005$ \\
0.75 & $+0.0011$ & $+0.0002$ \\
\midrule
\multicolumn{3}{l}{\textit{Reference (SDXL):} boost U0 $= +0.0061$ ($p{=}0.0168$, BH $q{=}0.0840$)} \\
\multicolumn{3}{l}{\textit{Reference (SDXL):} zeroing D2 $= +0.0054$ (BH $q{=}0.8273$)} \\
\bottomrule
\end{tabular}
\end{table}

\section{Compute Resources}
\label{app:compute}

All experiments were run on NVIDIA A100 (80GB) GPUs via SLURM. Table~\ref{tab:compute} breaks down approximate GPU-hours by experiment phase. The total (${\sim}159$ GPU-hours) includes all three analysis components (CSI pipeline, embedding repair, and per-block modality decomposition) across two primary models and one extension model.

\begin{table}[h]
\centering
\caption{Compute budget by experiment phase.}
\label{tab:compute}
\small
\begin{tabular}{lc}
\toprule
Experiment & GPU-hours (approx.) \\
\midrule
CSI pipeline (cross-model factorization, SD1.5+SDXL+SD3) & 24 \\
Embedding repair (pooled + token adapters, rank sweep) & 12 \\
Output-side BLIP scoring & 4 \\
Per-block boost sweep (432 $\times$ 5 seeds $\times$ 2 models) & 43 \\
Per-block zeroing sweep (432 $\times$ 3 seeds $\times$ 2 models) & 20 \\
Attr\_zero full-scale (432 $\times$ 5 seeds $\times$ 2 models) & 36 \\
Fractional attenuation (432 $\times$ 3 seeds $\times$ 4 levels) & 8 \\
Per-timestep phase analysis & 4 \\
Qualitative image generation & 1 \\
CLIP auxiliary subset & 7 \\
\midrule
\textbf{Total} & $\sim$159 \\
\bottomrule
\end{tabular}
\end{table}

\clearpage
\section*{NeurIPS Paper Checklist}

\begin{enumerate}

\item {\bf Claims}
    \item[] Question: Do the main claims made in the abstract and introduction accurately reflect the paper's contributions and scope?
    \item[] Answer: \answerYes{}
    \item[] Justification: The Abstract and \S\ref{sec:intro} state four contributions: (1) a diagnosis--control dissociation under the tested intervention families; (2) a modality-by-site decomposition showing different response profiles for selective token edits and broad cross-attention ablations; (3) text-only CSI as a pre-denoising stress coordinate across SD1.5, SDXL, and the SD3 text path; and (4) a matched six-prompt localization study connecting improved embedding fit under residual-minimizing adapters with CHR responses to downstream intervention. \S\ref{sec:limitations} distinguishes established from suggestive findings.
    \item[] Guidelines:
    \begin{itemize}
        \item The answer \answerNA{} means that the abstract and introduction do not include the claims made in the paper.
        \item The abstract and/or introduction should clearly state the claims made, including the contributions made in the paper and important assumptions and limitations. A \answerNo{} or \answerNA{} answer to this question will not be perceived well by the reviewers.
        \item The claims made should match theoretical and experimental results, and reflect how much the results can be expected to generalize to other settings.
        \item It is fine to include aspirational goals as motivation as long as it is clear that these goals are not attained by the paper.
    \end{itemize}

\item {\bf Limitations}
    \item[] Question: Does the paper discuss the limitations of the work performed by the authors?
    \item[] Answer: \answerYes{}
    \item[] Justification: \S\ref{sec:limitations} contains an explicit Established-vs-Suggestive split listing seven specific scope statements, followed by a key-caveats paragraph (small effect sizes, color-only endpoint, single SD3 path, two UNet architectures, fixed hyperparameters, untested broad-additive cell, untested WiCLP-aligned objective, fixed object mask).
    \item[] Guidelines:
    \begin{itemize}
        \item The answer \answerNA{} means that the paper has no limitation while the answer \answerNo{} means that the paper has limitations, but those are not discussed in the paper.
        \item The authors are encouraged to create a separate ``Limitations'' section in their paper.
        \item The paper should point out any strong assumptions and how robust the results are to violations of these assumptions (e.g., independence assumptions, noiseless settings, model well-specification, asymptotic approximations only holding locally). The authors should reflect on how these assumptions might be violated in practice and what the implications would be.
        \item The authors should reflect on the scope of the claims made, e.g., if the approach was only tested on a few datasets or with a few runs. In general, empirical results often depend on implicit assumptions, which should be articulated.
        \item The authors should reflect on the factors that influence the performance of the approach. For example, a facial recognition algorithm may perform poorly when image resolution is low or images are taken in low lighting. Or a speech-to-text system might not be used reliably to provide closed captions for online lectures because it fails to handle technical jargon.
        \item The authors should discuss the computational efficiency of the proposed algorithms and how they scale with dataset size.
        \item If applicable, the authors should discuss possible limitations of their approach to address problems of privacy and fairness.
        \item While the authors might fear that complete honesty about limitations might be used by reviewers as grounds for rejection, a worse outcome might be that reviewers discover limitations that aren't acknowledged in the paper. The authors should use their best judgment and recognize that individual actions in favor of transparency play an important role in developing norms that preserve the integrity of the community. Reviewers will be specifically instructed to not penalize honesty concerning limitations.
    \end{itemize}

\item {\bf Theory assumptions and proofs}
    \item[] Question: For each theoretical result, does the paper provide the full set of assumptions and a complete (and correct) proof?
    \item[] Answer: \answerNA{}
    \item[] Justification: This is an empirical mechanism-analysis paper. It contains no formal theorems or proofs; the only equations are operational definitions of CSI's non-factorization residual and DA, both of which are stated and used as measurement procedures rather than as objects of formal proof.
    \item[] Guidelines:
    \begin{itemize}
        \item The answer \answerNA{} means that the paper does not include theoretical results.
        \item All the theorems, formulas, and proofs in the paper should be numbered and cross-referenced.
        \item All assumptions should be clearly stated or referenced in the statement of any theorems.
        \item The proofs can either appear in the main paper or the supplemental material, but if they appear in the supplemental material, the authors are encouraged to provide a short proof sketch to provide intuition.
        \item Inversely, any informal proof provided in the core of the paper should be complemented by formal proofs provided in appendix or supplemental material.
        \item Theorems and Lemmas that the proof relies upon should be properly referenced.
    \end{itemize}

    \item {\bf Experimental result reproducibility}
    \item[] Question: Does the paper fully disclose all the information needed to reproduce the main experimental results of the paper to the extent that it affects the main claims and/or conclusions of the paper (regardless of whether the code and data are provided or not)?
    \item[] Answer: \answerYes{}
    \item[] Justification: \S\ref{sec:setup} specifies the stress-pair protocol (432 prompts, 216 paired groups), model scope, UNet block groups, diagnostic definitions, CHR endpoint, and statistical testing. Appendix~\ref{app:modality} states the denoiser settings: boost strength 4.0, mask quantile 0.80, 40 denoising timesteps, CFG scale 7.5, intervention throughout the full denoising trajectory, and five paired seeds for boost/\texttt{attr\_zero} versus three for broad zeroing. Appendices~\ref{app:csi}--\ref{app:compute} provide additional protocol details, per-block tables, prompt examples, and compute breakdown.
    \item[] Guidelines:
    \begin{itemize}
        \item The answer \answerNA{} means that the paper does not include experiments.
        \item If the paper includes experiments, a \answerNo{} answer to this question will not be perceived well by the reviewers: Making the paper reproducible is important, regardless of whether the code and data are provided or not.
        \item If the contribution is a dataset and\slash or model, the authors should describe the steps taken to make their results reproducible or verifiable.
        \item Depending on the contribution, reproducibility can be accomplished in various ways. For example, if the contribution is a novel architecture, describing the architecture fully might suffice, or if the contribution is a specific model and empirical evaluation, it may be necessary to either make it possible for others to replicate the model with the same dataset, or provide access to the model. In general. releasing code and data is often one good way to accomplish this, but reproducibility can also be provided via detailed instructions for how to replicate the results, access to a hosted model (e.g., in the case of a large language model), releasing of a model checkpoint, or other means that are appropriate to the research performed.
        \item While NeurIPS does not require releasing code, the conference does require all submissions to provide some reasonable avenue for reproducibility, which may depend on the nature of the contribution. For example
        \begin{enumerate}
            \item If the contribution is primarily a new algorithm, the paper should make it clear how to reproduce that algorithm.
            \item If the contribution is primarily a new model architecture, the paper should describe the architecture clearly and fully.
            \item If the contribution is a new model (e.g., a large language model), then there should either be a way to access this model for reproducing the results or a way to reproduce the model (e.g., with an open-source dataset or instructions for how to construct the dataset).
            \item We recognize that reproducibility may be tricky in some cases, in which case authors are welcome to describe the particular way they provide for reproducibility. In the case of closed-source models, it may be that access to the model is limited in some way (e.g., to registered users), but it should be possible for other researchers to have some path to reproducing or verifying the results.
        \end{enumerate}
    \end{itemize}

\item {\bf Open access to data and code}
    \item[] Question: Does the paper provide open access to the data and code, with sufficient instructions to faithfully reproduce the main experimental results, as described in supplemental material?
    \item[] Answer: \answerYes{}
    \item[] Justification: \S\ref{sec:conclusion} commits to releasing code, CSI computation scripts, prompts, and seeds on GitHub upon publication. The stress-prompt dataset is constructed from publicly available T2I-CompBench, CLEVR, and ARO lexicons; the diffusion backbones (SD1.5, SDXL, SD3) are public.
    \item[] Guidelines:
    \begin{itemize}
        \item The answer \answerNA{} means that paper does not include experiments requiring code.
        \item Please see the NeurIPS code and data submission guidelines (\url{https://neurips.cc/public/guides/CodeSubmissionPolicy}) for more details.
        \item While we encourage the release of code and data, we understand that this might not be possible, so \answerNo{} is an acceptable answer. Papers cannot be rejected simply for not including code, unless this is central to the contribution (e.g., for a new open-source benchmark).
        \item The instructions should contain the exact command and environment needed to run to reproduce the results. See the NeurIPS code and data submission guidelines (\url{https://neurips.cc/public/guides/CodeSubmissionPolicy}) for more details.
        \item The authors should provide instructions on data access and preparation, including how to access the raw data, preprocessed data, intermediate data, and generated data, etc.
        \item The authors should provide scripts to reproduce all experimental results for the new proposed method and baselines. If only a subset of experiments are reproducible, they should state which ones are omitted from the script and why.
        \item At submission time, to preserve anonymity, the authors should release anonymized versions (if applicable).
        \item Providing as much information as possible in supplemental material (appended to the paper) is recommended, but including URLs to data and code is permitted.
    \end{itemize}

\item {\bf Experimental setting/details}
    \item[] Question: Does the paper specify all the training and test details (e.g., data splits, hyperparameters, how they were chosen, type of optimizer) necessary to understand the results?
    \item[] Answer: \answerYes{}
    \item[] Justification: Appendix~\ref{app:modality} specifies intervention strength, mask quantiles, denoising steps, intervention schedule, CFG scale, and seed counts for the full per-block experiments. \S\ref{sec:setup} states the sign-flip and bootstrap draw counts and the FDR correction procedure. Adapter training details are given in Appendix~\ref{app:embed}, and the phase-restricted intervention analysis is described in Appendix~\ref{app:timestep}.
    \item[] Guidelines:
    \begin{itemize}
        \item The answer \answerNA{} means that the paper does not include experiments.
        \item The experimental setting should be presented in the core of the paper to a level of detail that is necessary to appreciate the results and make sense of them.
        \item The full details can be provided either with the code, in appendix, or as supplemental material.
    \end{itemize}

\item {\bf Experiment statistical significance}
    \item[] Question: Does the paper report error bars suitably and correctly defined or other appropriate information about the statistical significance of the experiments?
    \item[] Answer: \answerYes{}
    \item[] Justification: Figures~\ref{fig:da_ce}--\ref{fig:samesite} and Figures~\ref{fig:fragility} and~\ref{fig:polarity} report pointwise bootstrap-based 95\% confidence intervals; \S\ref{sec:results2} also reports confidence intervals for the localization results. The main CE table reports effect estimates and significance information, not confidence intervals. \S\ref{sec:setup} specifies the per-block two-sided sign-flip tests (5,000 Monte Carlo draws with plus-one correction; exact enumeration for at most 16 units), 2,000 bootstrap draws, and within-model/run BH correction. Appendix~\ref{app:csi} separately reports the group-preserving CSI permutation control. Per-prompt effect sizes ($d_z$) are reported in \S\ref{sec:results2} and Appendix~\ref{app:power}; the latter also reports post-hoc power analysis.
    \item[] Guidelines:
    \begin{itemize}
        \item The answer \answerNA{} means that the paper does not include experiments.
        \item The authors should answer \answerYes{} if the results are accompanied by error bars, confidence intervals, or statistical significance tests, at least for the experiments that support the main claims of the paper.
        \item The factors of variability that the error bars are capturing should be clearly stated (for example, train/test split, initialization, random drawing of some parameter, or overall run with given experimental conditions).
        \item The method for calculating the error bars should be explained (closed form formula, call to a library function, bootstrap, etc.)
        \item The assumptions made should be given (e.g., Normally distributed errors).
        \item It should be clear whether the error bar is the standard deviation or the standard error of the mean.
        \item It is OK to report 1-sigma error bars, but one should state it. The authors should preferably report a 2-sigma error bar than state that they have a 96\% CI, if the hypothesis of Normality of errors is not verified.
        \item For asymmetric distributions, the authors should be careful not to show in tables or figures symmetric error bars that would yield results that are out of range (e.g., negative error rates).
        \item If error bars are reported in tables or plots, the authors should explain in the text how they were calculated and reference the corresponding figures or tables in the text.
    \end{itemize}

\item {\bf Experiments compute resources}
    \item[] Question: For each experiment, does the paper provide sufficient information on the computer resources (type of compute workers, memory, time of execution) needed to reproduce the experiments?
    \item[] Answer: \answerYes{}
    \item[] Justification: Appendix~\ref{app:compute} provides a per-experiment GPU-hour breakdown (CSI pipeline 24h, embedding repair 12h, BLIP scoring 4h, per-block boost sweep 43h, zeroing sweep 20h, attr\_zero full-scale 36h, fractional attenuation 8h, per-timestep analysis 4h, qualitative generation 1h, CLIP auxiliary 7h; total $\sim$159 GPU-hours on NVIDIA A100 80GB).
    \item[] Guidelines:
    \begin{itemize}
        \item The answer \answerNA{} means that the paper does not include experiments.
        \item The paper should indicate the type of compute workers CPU or GPU, internal cluster, or cloud provider, including relevant memory and storage.
        \item The paper should provide the amount of compute required for each of the individual experimental runs as well as estimate the total compute.
        \item The paper should disclose whether the full research project required more compute than the experiments reported in the paper (e.g., preliminary or failed experiments that didn't make it into the paper).
    \end{itemize}

\item {\bf Code of ethics}
    \item[] Question: Does the research conducted in the paper conform, in every respect, with the NeurIPS Code of Ethics \url{https://neurips.cc/public/EthicsGuidelines}?
    \item[] Answer: \answerYes{}
    \item[] Justification: The work is purely an analysis of existing publicly available diffusion models on synthetic prompts. No human subjects, no scraping, no new generative model, no deployment. The research conforms with the NeurIPS Code of Ethics.
    \item[] Guidelines:
    \begin{itemize}
        \item The answer \answerNA{} means that the authors have not reviewed the NeurIPS Code of Ethics.
        \item If the authors answer \answerNo, they should explain the special circumstances that require a deviation from the Code of Ethics.
        \item The authors should make sure to preserve anonymity (e.g., if there is a special consideration due to laws or regulations in their jurisdiction).
    \end{itemize}

\item {\bf Broader impacts}
    \item[] Question: Does the paper discuss both potential positive societal impacts and negative societal impacts of the work performed?
    \item[] Answer: \answerNA{}
    \item[] Justification: This is an analysis paper that does not introduce a new generative model or any new capability that did not previously exist. The interventions we study are diagnostic and per-block ablations of existing public diffusion models; they do not increase the misuse surface of those models. Any improvement in compositional fidelity from following our diagnostic guidance applies equally to legitimate and illegitimate uses of text-to-image generation, but does not introduce a new vector of misuse.
    \item[] Guidelines:
    \begin{itemize}
        \item The answer \answerNA{} means that there is no societal impact of the work performed.
        \item If the authors answer \answerNA{} or \answerNo, they should explain why their work has no societal impact or why the paper does not address societal impact.
        \item Examples of negative societal impacts include potential malicious or unintended uses (e.g., disinformation, generating fake profiles, surveillance), fairness considerations (e.g., deployment of technologies that could make decisions that unfairly impact specific groups), privacy considerations, and security considerations.
        \item The conference expects that many papers will be foundational research and not tied to particular applications, let alone deployments. However, if there is a direct path to any negative applications, the authors should point it out. For example, it is legitimate to point out that an improvement in the quality of generative models could be used to generate Deepfakes for disinformation. On the other hand, it is not needed to point out that a generic algorithm for optimizing neural networks could enable people to train models that generate Deepfakes faster.
        \item The authors should consider possible harms that could arise when the technology is being used as intended and functioning correctly, harms that could arise when the technology is being used as intended but gives incorrect results, and harms following from (intentional or unintentional) misuse of the technology.
        \item If there are negative societal impacts, the authors could also discuss possible mitigation strategies (e.g., gated release of models, providing defenses in addition to attacks, mechanisms for monitoring misuse, mechanisms to monitor how a system learns from feedback over time, improving the efficiency and accessibility of ML).
    \end{itemize}

\item {\bf Safeguards}
    \item[] Question: Does the paper describe safeguards that have been put in place for responsible release of data or models that have a high risk for misuse (e.g., pre-trained language models, image generators, or scraped datasets)?
    \item[] Answer: \answerNA{}
    \item[] Justification: We release no new generative model and no scraped dataset. The released artifacts are analysis code and a synthetic stress-pair prompt list constructed from existing public benchmarks.
    \item[] Guidelines:
    \begin{itemize}
        \item The answer \answerNA{} means that the paper poses no such risks.
        \item Released models that have a high risk for misuse or dual-use should be released with necessary safeguards to allow for controlled use of the model, for example by requiring that users adhere to usage guidelines or restrictions to access the model or implementing safety filters.
        \item Datasets that have been scraped from the Internet could pose safety risks. The authors should describe how they avoided releasing unsafe images.
        \item We recognize that providing effective safeguards is challenging, and many papers do not require this, but we encourage authors to take this into account and make a best faith effort.
    \end{itemize}

\item {\bf Licenses for existing assets}
    \item[] Question: Are the creators or original owners of assets (e.g., code, data, models), used in the paper, properly credited and are the license and terms of use explicitly mentioned and properly respected?
    \item[] Answer: \answerYes{}
    \item[] Justification: The diffusion models and source benchmarks (T2I-CompBench, CLEVR, and ARO) are credited in \S\ref{sec:setup} and \S\ref{sec:related}. The model licenses are distinct: SD1.5 uses the \href{https://huggingface.co/stable-diffusion-v1-5/stable-diffusion-v1-5/blob/main/LICENSE}{CreativeML Open RAIL-M License}; SDXL Base 1.0 uses the \href{https://huggingface.co/stabilityai/stable-diffusion-xl-base-1.0/blob/main/LICENSE.md}{CreativeML Open RAIL++-M License}; and the SD3 Medium Diffusers repository used for the text-path extension specifies the \href{https://huggingface.co/stabilityai/stable-diffusion-3-medium-diffusers/blob/main/LICENSE}{Stability AI Non-Commercial Research Community License}. Benchmark assets remain subject to their respective authors' terms. These assets are used for the research analyses described in the paper.
    \item[] Guidelines:
    \begin{itemize}
        \item The answer \answerNA{} means that the paper does not use existing assets.
        \item The authors should cite the original paper that produced the code package or dataset.
        \item The authors should state which version of the asset is used and, if possible, include a URL.
        \item The name of the license (e.g., CC-BY 4.0) should be included for each asset.
        \item For scraped data from a particular source (e.g., website), the copyright and terms of service of that source should be provided.
        \item If assets are released, the license, copyright information, and terms of use in the package should be provided. For popular datasets, \url{paperswithcode.com/datasets} has curated licenses for some datasets. Their licensing guide can help determine the license of a dataset.
        \item For existing datasets that are re-packaged, both the original license and the license of the derived asset (if it has changed) should be provided.
        \item If this information is not available online, the authors are encouraged to reach out to the asset's creators.
    \end{itemize}

\item {\bf New assets}
    \item[] Question: Are new assets introduced in the paper well documented and is the documentation provided alongside the assets?
    \item[] Answer: \answerYes{}
    \item[] Justification: The new assets are analysis code and a stress-pair prompt list. Both will be released on GitHub upon publication (\S\ref{sec:conclusion}) with documentation on protocol, prompt construction, intervention hyperparameters, seeds, and analysis scripts. Prompt examples appear in Appendix~\ref{app:prompts}.
    \item[] Guidelines:
    \begin{itemize}
        \item The answer \answerNA{} means that the paper does not release new assets.
        \item Researchers should communicate the details of the dataset\slash code\slash model as part of their submissions via structured templates. This includes details about training, license, limitations, etc.
        \item The paper should discuss whether and how consent was obtained from people whose asset is used.
        \item At submission time, remember to anonymize your assets (if applicable). You can either create an anonymized URL or include an anonymized zip file.
    \end{itemize}

\item {\bf Crowdsourcing and research with human subjects}
    \item[] Question: For crowdsourcing experiments and research with human subjects, does the paper include the full text of instructions given to participants and screenshots, if applicable, as well as details about compensation (if any)?
    \item[] Answer: \answerNA{}
    \item[] Justification: The paper does not involve crowdsourcing or human-subjects research. All evaluation uses automated metrics (object-local color hit rate, BLIP-VQA scores, CLIP similarity).
    \item[] Guidelines:
    \begin{itemize}
        \item The answer \answerNA{} means that the paper does not involve crowdsourcing nor research with human subjects.
        \item Including this information in the supplemental material is fine, but if the main contribution of the paper involves human subjects, then as much detail as possible should be included in the main paper.
        \item According to the NeurIPS Code of Ethics, workers involved in data collection, curation, or other labor should be paid at least the minimum wage in the country of the data collector.
    \end{itemize}

\item {\bf Institutional review board (IRB) approvals or equivalent for research with human subjects}
    \item[] Question: Does the paper describe potential risks incurred by study participants, whether such risks were disclosed to the subjects, and whether Institutional Review Board (IRB) approvals (or an equivalent approval/review based on the requirements of your country or institution) were obtained?
    \item[] Answer: \answerNA{}
    \item[] Justification: No human subjects research; no IRB approval required.
    \item[] Guidelines:
    \begin{itemize}
        \item The answer \answerNA{} means that the paper does not involve crowdsourcing nor research with human subjects.
        \item Depending on the country in which research is conducted, IRB approval (or equivalent) may be required for any human subjects research. If you obtained IRB approval, you should clearly state this in the paper.
        \item We recognize that the procedures for this may vary significantly between institutions and locations, and we expect authors to adhere to the NeurIPS Code of Ethics and the guidelines for their institution.
        \item For initial submissions, do not include any information that would break anonymity (if applicable), such as the institution conducting the review.
    \end{itemize}

\item {\bf Declaration of LLM usage}
    \item[] Question: Does the paper describe the usage of LLMs if it is an important, original, or non-standard component of the core methods in this research? Note that if the LLM is used only for writing, editing, or formatting purposes and does \emph{not} impact the core methodology, scientific rigor, or originality of the research, declaration is not required.
    \item[] Answer: \answerNA{}
    \item[] Justification: LLMs are not part of the core methodology. The core method involves analysis of cross-attention in diffusion U-Nets and non-factorization residuals in CLIP-family text encoders.
    \item[] Guidelines:
    \begin{itemize}
        \item The answer \answerNA{} means that the core method development in this research does not involve LLMs as any important, original, or non-standard components.
        \item Please refer to our LLM policy in the NeurIPS handbook for what should or should not be described.
    \end{itemize}

\end{enumerate}

\end{document}